\documentclass{article}
\usepackage[utf8]{inputenc}

\usepackage{microtype}
\usepackage{graphicx}
\usepackage{subfigure}
\usepackage{booktabs} 
\usepackage{colortbl}
\usepackage{courier} 
\usepackage{listings}
\usepackage{listingsutf8}
\usepackage[most]{tcolorbox} 
\usepackage{multirow}   
\usepackage{hyperref}

\usepackage{multirow}

\usepackage{hyperref}

\usepackage[accepted]{icml2025}

\usepackage{amsmath}
\usepackage{amssymb}
\usepackage{mathtools}
\usepackage{amsthm}

\usepackage{hyperref}
\usepackage{xcolor}
\hypersetup{
    colorlinks=true,
    linkcolor=blue,
    filecolor=magenta,      
    urlcolor=teal, 
}

\usepackage[capitalize,noabbrev]{cleveref}

\theoremstyle{plain}
\newtheorem{theorem}{Theorem}[section]

\theoremstyle{definition}

\newtheorem{assumption}[theorem]{Assumption}
\theoremstyle{remark}

\usepackage[textsize=tiny]{todonotes}

\icmltitlerunning{Yunjue Agent Tech Report}

\begin{document}

\twocolumn[
\icmltitle{Yunjue Agent Tech Report: A Fully Reproducible, Zero-Start \\ In-Situ Self-Evolving Agent System for Open-Ended Tasks}

\icmlsetsymbol{equal}{*}
\icmlsetsymbol{InternYunjue}{$\dagger$}

\begin{icmlauthorlist}
\icmlauthor{Haotian Li}{equal,InternYunjue,YunjueTech,hagongda}
\icmlauthor{Shijun Yang}{equal,InternYunjue,USTC,YunjueTech}
\icmlauthor{Weizhen Qi}{YunjueTech}
\icmlauthor{Silei Zhao}{YunjueTech} \\
\icmlauthor{Rui Hua}{YunjueTech}
\icmlauthor{Mingzhu Song}{YunjueTech}
\icmlauthor{Xiaojian Yang}{YunjueTech}
\icmlauthor{Chao Peng}{YunjueTech}
\end{icmlauthorlist}

\icmlaffiliation{YunjueTech}{Yunjue Technology}
\icmlaffiliation{hagongda}{Harbin Institute of Technology}
\icmlaffiliation{USTC}{University of Science and Technology of China}

\begin{center}
    \vspace{0.1em} 
    \large 
    \textbf{Links:}
    \href{https://github.com/YunjueTech/Yunjue-Agent}{\texttt{[GitHub Repository]}} 
    \href{https://huggingface.co/datasets/YunjueTech/Yunjue-Agent-Traces}{\texttt{[System Traces]}}
    \vspace{1em}
\end{center}

\icmlcorrespondingauthor{Weizhen Qi, Tech Lead. }{qiweizhen@yunjuetech.com}

\icmlkeywords{Machine Learning, ICML}

\vskip 0.3in
]

\printAffiliationsAndNotice{\icmlEqualContribution\ $^{\dagger}$During Internship at Yunjue Technology} 
\begin{abstract}
Conventional agent systems often struggle in open-ended environments where task distributions continuously drift and external supervision is scarce. Their reliance on static toolsets or offline training renders capability boundaries rigid, preventing adaptation to evolving dynamics. To address this, we propose the \textit{In-situ Self-Evolving} paradigm. This approach treats sequential task interactions as a continuous stream of experience, enabling the system to distill short-term execution feedback into long-term, reusable capabilities without ground-truth labels. We identify tool evolution as the critical pathway for this capability expansion. Accordingly, we develop the \textit{Yunjue Agent}, a system that iteratively synthesizes, optimizes, and reuses tools to navigate emerging challenges. To optimize evolutionary efficiency, we introduce a \textit{Parallel Batch Evolution} strategy. Furthermore, we propose a novel metric to monitor evolutionary convergence, serving as a function analogous to training loss in conventional optimization. Empirical evaluations across five diverse benchmarks under a zero-start setting demonstrate significant performance gains over proprietary baselines and validate the effectiveness of our proposed convergence metric. Additionally, complementary warm-start evaluations confirm that the accumulated general knowledge can be seamlessly transferred to novel domains. We open-source our codebase, system traces, and evolved tools to facilitate future research in resilient, self-evolving intelligence.
\end{abstract}

\section{Introduction}
The rapid advancement of Large Language Models (LLMs) has catalyzed a paradigm shift in the pursuit of Artificial General Intelligence (AGI)~\cite{10.24963/ijcai.2024/890}, fundamentally expanding the AI frontier across diverse daily and scientific domains. While these developments have birthed versatile agents, a critical limitation persists: most contemporary systems remain confined within predefined operational envelopes. This reliance on static functional boundaries renders agents inherently fragile when confronted with the stochasticity of open-world environments. We posit that true AGI demands not merely broad-domain competence, but the capacity for self-adaptive evolution—dynamically reshaping capabilities to surmount novel challenges within open-ended contexts. Consequently, architecting a system capable of both cross-domain execution and recursive self-evolution has emerged as a paramount challenge in the quest for resilient intelligence.

Within the framework of LLM-based agents, such adaptive capability typically rests on three pillars: \textit{workflow, context, and tool}. Recently, the field has witnessed a surge in \textit{self-evolving agents}~\cite{tao2024survey,gao2025survey}, designed to autonomously enhance capabilities without human intervention, achieving progress in workflow adaptation~\cite{zhang2025aflow}, context management~\cite{tang2025agent}, and tool utilization~\cite{wang2023voyager}. However, existing paradigms often rely on offline supervision~\cite{acikgoz2025self} or remain confined to narrow domains~\cite{wang2023voyager,jin2025stella}. In supervision-scarce, open-ended environments, a more autonomous paradigm is imperative. We argue that among the three pillars of agent systems, while workflows refine expertise and context aligns preference, the toolset constitutes the fundamental prerequisite for novel environments. Without requisite functional tools, even sophisticated planning or memory becomes computationally paralyzed. Furthermore, tool functionality is verifiable via rigorous execution: a program either succeeds or throws an exception. This robust binary signal enables autonomous self-correction even absent human intervention, positioning tools as the optimal starting point for in-situ evolution. 

To address the open-ended challenges, we propose the \textbf{In-situ Self-Evolving (ISE)} framework. This paradigm treats task interactions as a continuous stream for accumulating generalized knowledge, initiating with a \textit{zero-start} phase. We develop the \textbf{Yunjue Agent}, utilizing a multi-agent architecture (Manager, Tool Developer, Executor, etc.)~\cite{qian2024scaling} to synthesize bespoke Python primitives on-the-fly when existing capabilities are insufficient. To further optimize efficiency, we introduce a \textit{Parallel Batch Evolution} strategy, which distills high-quality tools from concurrent executions to accelerate convergence. Furthermore, we propose a novel metric to monitor evolutionary convergence, functioning analogously to training loss. We open-source our codebase, complete interaction traces, and the library of evolved tools to provide a transparent foundation for future research in resilient, self-evolving intelligence.

Comprehensive empirical evaluations across five benchmarks validate the Yunjue Agent's efficacy. Results demonstrate: (i) \textbf{State-of-the-art performance}: In zero-start settings, autonomous tool evolution outperforms proprietary baselines; (ii) \textbf{Cross-domain transferability}: Additional warm-start evaluations show that accumulated general knowledge adapts effectively to novel domains; (iii) \textbf{Robust convergence monitoring}: Our proposed metric effectively serves as a stability proxy for evolution; and (iv) \textbf{Cost-effectiveness}: The lightweight design ensures economical operation.

\section{In-situ self-evolving agents}
\textbf{An agent system} can be formally defined as a tuple $\mathcal{M} = \langle \mathcal{W}, \mathcal{C}, \mathcal{T} \rangle$ of workflow, context and tools \cite{gao2025survey}. In this system, the workflow $\mathcal{W}$ is typically structured as a directed graph $\mathcal{G} = (V, E)$, where nodes $V$ represent LLM-based agents and edges $E$ denote the flow of information. $\mathcal{C}$ is the set of contexts, where each context element can be a prompt template or a memory buffer encompassing dynamic conversation history. Finally, $\mathcal{T}$ denotes the set of tools available for the agents to execute specific tasks.

To enable agents to adapt to new environments like humans, recent work focuses on self-evolving methods~\cite{tao2024survey,gao2025survey}. However, these methods typically require an iterative training process to update certain components~\cite{he2025visplay,chen2025mathse} or are confined in specific areas \cite{xia2025live,jin2025stella}, such as generating query-specific workflows~\cite{zhang2025aflow,ye2025masgpt}, refining context $\mathcal{C}$ (e.g., prompt and memory)~\cite{zhang2025aflow,tang2025agent} or adjusting toolset~\cite{wang2024agent, wang2023voyager}. In contrast, we aim to explore whether an agent system can adapt to the environment by continuously updating its components in-situ, where there is no access to external supervision signals. This motivation leads to a new paradigm of agentic evolution.

\textbf{In-situ self-evolving of an agent system}. Given a sequence of queries $\{x_1, x_2, \dots, x_T\}$, the agent evolves dynamically from $\mathcal{M}_0$ to $\mathcal{M}_T$ as it processes each. Specifically, after completing the $t$-th query $x_t$, the system updates its configuration for the next query, which can be formalized as a transition from $\mathcal{M}_{t-1} = \langle \mathcal{W}_{t-1}, \mathcal{C}_{t-1}, \mathcal{T}_{t-1} \rangle$ to $\mathcal{M}_{t} = \langle \mathcal{W}_{t}, \mathcal{C}_{t}, \mathcal{T}_{t} \rangle$, where any component of the tuple—the workflow structure, context, or toolset—may be modified based on internal feedback or experience gained from the previous interaction. Unlike self-evolving agents that maximize a specific objective function through a series of training steps \citep{gao2025survey}, \textit{in-situ self evolving} operates during the inference phase where ground truth information is unavailable.

To ensure a focused scope, we fix the workflow $\mathcal{W}$. Furthermore, given that Agent KB \citep{tang2025agent} has attempted to optimize memory through cross-domain experience to enhance agent performance, we also treat the context $\mathcal{C}$ as a fixed component\footnote{Although the execution flow can be dynamic during inference due to branch control and model inputs are populated at runtime, the underlying workflow structure and prompt templates are pre-defined. Thus, we consider these configurations fixed.}, reducing the evolving system state to $\mathcal{M}_t = \langle \mathcal{W}_0, \mathcal{C}_0, \mathcal{T}_t \rangle$. We posit that the variability of the toolset is the most decisive factor for a general-purpose agent system. While memory enhances performance by recalling experience, the ability to effectively solve problems in novel domains is fundamentally limited by the availability of appropriate tools~\citep{wang2023voyager, qin2024toolllm}.

\begin{figure*}[t]
    \centering
    \includegraphics[width=\linewidth]{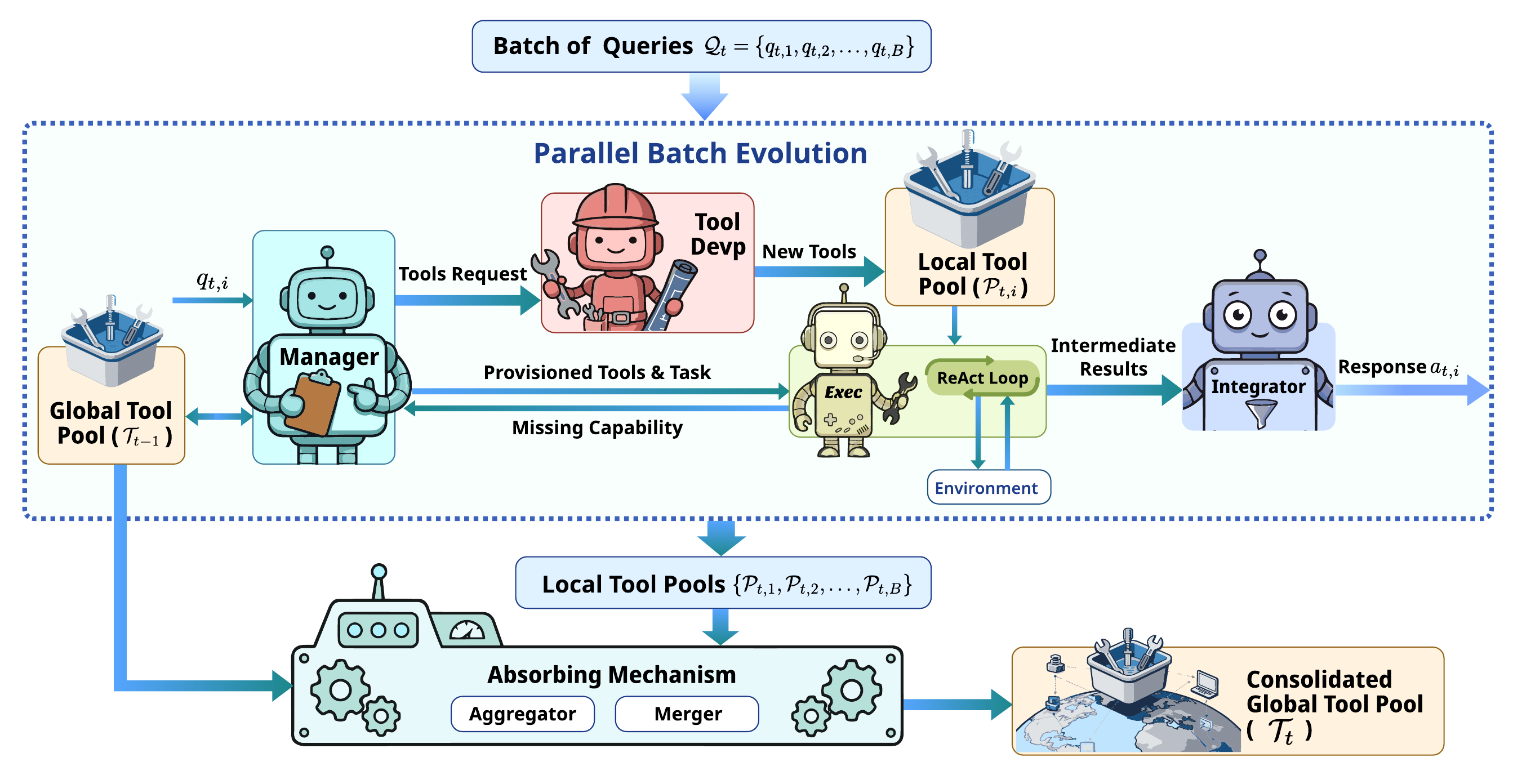}
    \vspace*{-.6cm}
    \caption{An architecture overview of Yunjue Agent.}\label{fig:overview}
    \vspace*{-.3cm}
\end{figure*}

\section{Methodology}
\subsection{In-situ self-evolving via tool accumulation}
We posit that the fundamental prerequisite for a generalist agent lies in the continuous, dynamic evolution of its toolset. To address this, we introduce an agentic workflow for \textit{in-situ self-evolving} via tool accumulation.

Upon receiving a query $x_t$, the agent first attempts to retrieve relevant utilities from its existing repository, denoted as $\mathcal{T}_{sub} \subseteq \mathcal{T}_{t-1}$. In the absence of requisite capabilities, the agent synthesizes novel tools $\mathcal{P}_{t}$ tailored to the specific task constraints. Subsequently, empowered by the augmented toolset $\mathcal{T}_{sub} \cup \mathcal{P}_{t}$, the agent executes the task through a sequence of tool invocations. To ensure that generated tools are not merely functional but also robust and reusable, the agent engages in a self-reflection mechanism post-execution, refining the tools based on the error reports and execution traces. The process concludes with the integration of these newly synthesized tools into the global repository, updating the state to $\mathcal{T}_{t} = \mathcal{P}_{t} \cup \mathcal{T}_{t-1}$.

As the agent processes an expanding stream of queries, existing tools undergo iterative refinement while new ones are concurrently synthesized. This dual mechanism propels the system's evolution along two dimensions: \textit{breadth} (expanding functional coverage) and \textit{depth} (optimizing tool robustness). Consequently, upon exposure to a sufficient volume of tasks, the tool repository is expected to reach a state of convergence, where incremental synthesis is necessitated only by outlier queries with idiosyncratic requirements.

\subsection{Parallel batch evolution}\label{sec:parallel_batch_evolution}
To balance processing efficiency with shared knowledge accumulation, we propose a \textit{Parallel Batch Evolution} strategy. While sequential processing maximizes reuse, it lacks scalability; conversely, independent parallelization sacrifices shared learning. Our approach reconciles these by parallelizing execution while maintaining a cohesive evolutionary trajectory.

Formally, let $\mathcal{Q}_t = \{q_{t,1}, q_{t,2}, \dots, q_{t,B}\}$ denote a batch of $B$ user queries input to the agent $\mathcal{M}_{t-1}$ at step $t$. For each query $q_{t,i} \in \mathcal{Q}_t$, the system synthesizes a set of local tools $\mathcal{P}_{t,i}$ that augments the global toolset $\mathcal{T}_{t-1}$ specifically for that instance. However, independent generation often yields functionally redundant tools across different queries—particularly for general-purpose utilities like web searching. This redundancy expands the tool search space, increasing the cognitive load on the agent. To mitigate this, we introduce a \emph{tool absorbing mechanism} designed to cluster and consolidate utilities post-generation. Specifically, upon completion of batch $\mathcal{Q}_t$, the system aggregates all tools $\{\mathcal{T}_{t-1}, \mathcal{P}_{t,1}, \dots, \mathcal{P}_{t,B}\}$ and clusters them into disjoint groups $\{G_j\}$ based on functional semantic similarity. Subsequently, a merging function $\Phi$ consolidates these groups, filtering for quality and redundancy to produce a compact updated pool $\mathcal{T}_{t} = \Phi(\{G_j\})$. This process ensures the tool space remains streamlined, preventing retrieval ambiguity while updating the system state to $\mathcal{M}_{t}$.

Processing queries in batches offers distinct advantages. Primarily, it significantly enhances system throughput. Furthermore, analogous to mini-batch gradient descent in model optimization~\citep{bottou2010large, goodfellow2016deep}—which mitigates gradient variance by averaging over samples—our absorbing mechanism reduces evolutionary stochasticity by merging similar tool instances. Simultaneously, this acts as a form of \textit{Best-of-N} test-time scaling~\citep{cobbe2021training, beirami2024theoretical}, effectively performing multiple parallel rollouts for tool creation and selecting the optimal synthesis results for the permanent library.

\subsection{The Yunjue Agent system}
By decoupling tool management, synthesis, and task execution, we establish a multi-agent system~\citep{10.1145/3586183.3606763} optimized for \textit{in-situ self-evolving}. As illustrated in Figure~\ref{fig:overview}, the architecture comprises distinct functional roles: \textbf{Manager}, \textbf{Executor}, \textbf{Tool Developer}, and \textbf{Integrator}, supported by the \textbf{Aggregator} and \textbf{Merger} for batch-level synchronization. The workflow proceeds via a collaborative mechanism designed to handle complex queries flexibly:

Upon receiving a user query $q_t$, the \textit{Manager} orchestrates the workflow by first aligning tool capabilities with task requirements. It analyzes the system state to retrieve a relevant subset of tools $\mathcal{T}_{\text{sub}} \subseteq \mathcal{T}_{t-1}$ from the global repository. Should capability gaps arise, the \textit{Manager} directs the \textit{Tool Developer} to synthesize bespoke tools (implemented as Python primitives), which are immediately instantiated in the local context. Subsequently, the \textit{Executor} addresses the query using the provisioned toolset, adhering to the ReAct paradigm~\cite{yao2023react}. Crucially, the system supports dynamic runtime adaptation: if the \textit{Executor} encounters unforeseen capability deficits during reasoning (e.g., identifying the need for a specific file parser), it suspends execution and signals the \textit{Manager}. The \textit{Manager} then provisions the requisite tools on-the-fly, allowing the \textit{Executor} to resume seamlessly.

\begin{figure*}[t]
    \centering
    \includegraphics[width=\linewidth]{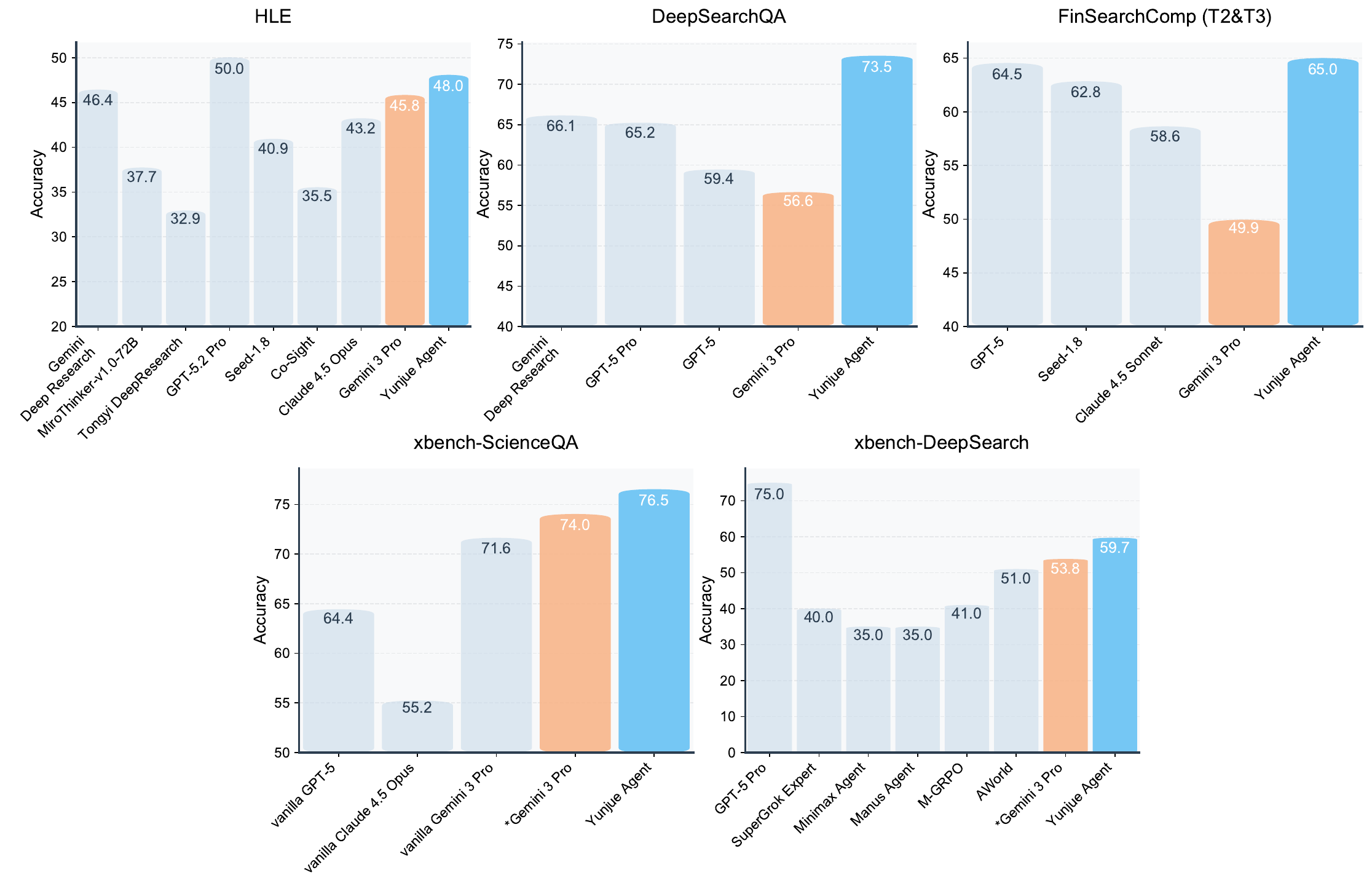}
    \vspace*{-0.7cm}
    \caption{Performance comparison of Yunjue Agent against state-of-the-art agents and agentic foundation models. Our method is highlighted in cyan, and the backend model (Gemini 3 Pro) appears in orange. \emph{*Gemini 3 Pro} denotes our implementation with a Python interpreter.}\label{fig:main_results}
    \vspace*{-0.4cm}
\end{figure*}

Upon task completion, the \textit{Integrator} consolidates the execution history and intermediate outputs to formulate a comprehensive response $a_t$. Through this iterative process, the Yunjue Agent evolves from state $\mathcal{M}_{t-1}$ to $\mathcal{M}_t$, effectively updating the tool pool to $\mathcal{T}_t$ for future evolution.

\textbf{Parallel evolution with batch processing.} To optimize evolutionary throughput, we implement the \emph{parallel batch evolution} strategy. Given a query batch $\mathcal{Q}_t = \{q_{t,1}, q_{t,2}, \dots, q_{t,B}\}$, the system processes each instance concurrently, granting all agents shared access to the global tool repository $\mathcal{T}_{t-1}$. This parallel execution allows each query $q_{t,i}$ to independently synthesize a local toolset $\mathcal{P}_{t,i}$ tailored to its specific execution context.

Upon the completion of the batch, the \emph{tool absorbing mechanism} is invoked to consolidate the dispersed local pools with the global state. Specifically, the system aggregates the union of all toolsets $\{\mathcal{T}_{t-1}, \mathcal{P}_{t,1}, \mathcal{P}_{t,2}, \dots, \mathcal{P}_{t,B}\}$. The LLM-based \textit{Aggregator} then groups utilities based on functional semantic similarity. Subsequently, another LLM-based \textit{Merger} is applied to each cluster to synthesize a unified, canonical tool that encapsulates collective capabilities while eliminating redundancy. The resulting consolidated repository $\mathcal{T}_t$ serves as the initialized state for the subsequent queries. The pseudo-code of the overall process is provided in Appendix~\ref{app:algorithm}.

\section{Experiment setup}
\textbf{Datasets}. To demonstrate the generalizability of our approach across diverse domains and task complexities, we conduct comprehensive evaluations on five complementary benchmarks, each targeting distinct professional scenarios: (i) \emph{HLE} (Humanity's Last Exam)~\cite{hle}, a frontier multi-modal benchmark featuring expert-level questions across mathematics, humanities, and natural sciences, designed to assess advanced reasoning at the boundary of human knowledge; (ii) \emph{DeepSearchQA} (DSQA)~\cite{deepsearchqa}, which challenges agents' ability to synthesize comprehensive answers through deep web search, iterative information gathering, and multi-source evidence aggregation; (iii) \emph{xBench}~\cite{xbench}, a Chinese profession-aligned evaluation suite where we focus on the \emph{ScienceQA} (xSciQA) subset (spanning natural, social, and language sciences) and \emph{DeepSearch} (xDS, 2025.10 version) subset to evaluate real-world productivity in scientific research and complex retrieval workflows, allowing us to assess cross-lingual adaptation capabilities; and (iv) \emph{FinSearchComp} (FSC)~\cite{finsearchcomp}, a bilingual (English and Chinese) benchmark targeting financial analysis through its \emph{T2} (Simple Historical Lookup) and \emph{T3} (Complex Historical Investigation) tasks, which require precise time-sensitive data retrieval and multi-step quantitative reasoning over financial documents.

\textbf{Baselines.} We benchmark our agent system against a comprehensive suite of proprietary and open-source systems, spanning both static and self-evolving agent paradigms. Unless otherwise noted, reported performance metrics are derived directly from original publications, technical reports, or authoritative disclosures (detailed provenance is provided in the Appendix~\ref{app:setup}).

\begin{figure*}[t]
    \centering
    \includegraphics[width=\linewidth]{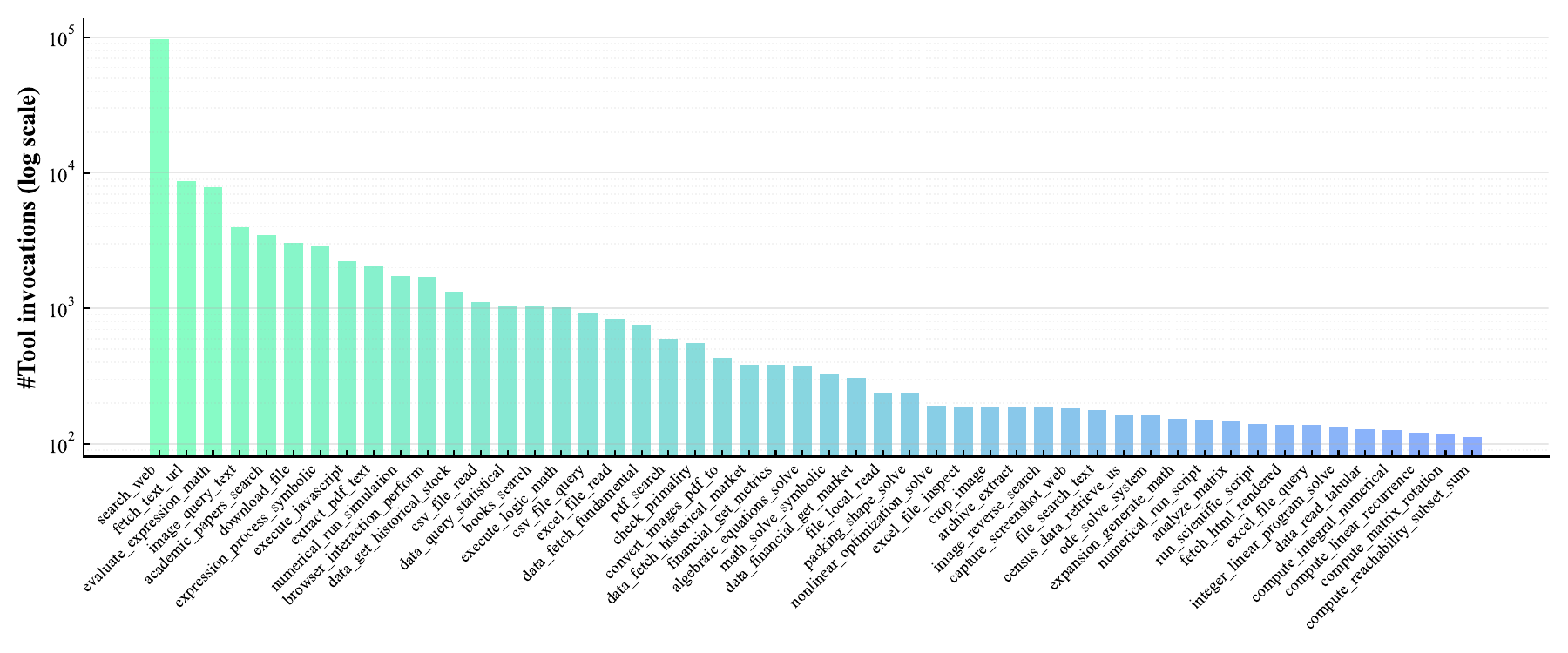}
    \vspace*{-0.7cm}
    \caption{Invocation frequency distribution of the toolset evolved across five benchmarks. We report the top 50 tools, illustrating the emergence of high-generalizability primitives.}\label{fig:main_results_tool_usage}
    \vspace*{-0.45cm}
\end{figure*}

\section{Evaluation on open-ended evolution}
\subsection{Zero-start performance on cross-domain tasks}
We evaluate our Yunjue Agent, initialized with an empty toolset, across all five benchmarks against state-of-the-art baselines. Results are presented in Figure~\ref{fig:main_results}. For datasets excluding xSciQA, foundation model baselines (e.g., GPT-5, Gemini 3 Pro) are augmented with web search and Python interpreters. Conversely, on xSciQA, where standard leaderboards prohibit external tools, we explicitly equip our backend model with a Python interpreter to ensure a fair parity comparison. We provide more detailed results in Appendix~\ref{app:results}, including the standard deviation of accuracy, token consumption and cost.

On the heterogeneous HLE benchmark, our agent yields a significant improvement over the backend (48.0 vs. 45.8), ranking second only to GPT-5.2 Pro. Most notably, Yunjue Agent achieves state-of-the-art performance on DSQA, FSC, and xSciQA. We observe substantial absolute gains over the Gemini 3 Pro baseline, with increases of \textbf{+16.9} points on DSQA (73.5 vs. 56.6) and \textbf{+15.1} points on FSC (65.0 vs. 49.9), alongside a record-setting score of 76.5 on xSciQA. Finally, on xDS, our method maintains competitive performance (59.7), surpassed only by GPT-5 Pro while outperforming all other models by a significant margin. These exceptional results demonstrate our method's \textbf{superior performance across a wide spectrum of tasks, laying a solid foundation for generalized knowledge accumulation and transfer.}

Complementing the performance analysis, we aggregated toolsets across all five benchmarks to examine functional utilization patterns. Figure~\ref{fig:main_results_tool_usage} presents the invocation frequency of the top 50 tools. The distribution reveals the \textbf{spontaneous emergence} of high-utility fundamental functions, most notably \texttt{web\_search}, \texttt{fetch\_text\_url} and \texttt{evaluate\_expression\_math}. The dominance of these foundational tools confirms that the system has effectively \textbf{distilled generalized knowledge into versatile primitives}, ensuring broad applicability across diverse task semantics.

Although the Yunjue Agent integrates modules such as the \textit{Manager} and \textit{Tool Developer} around the \textit{Executor}, the \textit{Executor} acts as the primary entity for execution and token consumption. In Appendix~\ref{app:results}, we show that across five datasets, the \textit{Executor} accounts for approximately 95\% of the total token consumption. Additionally, since tools do not need to be repeatedly created, a significant amount of token consumption is also saved. Consequently, our system's overall overhead is comparable to that of a standard ReAct agent equipped with a Python interpreter, a finding further corroborated in Table~\ref{tab:ablation_tool_evolution}.

\subsection{Warm-start evolution in shifting domains}
We validate the transferability of accumulated knowledge to novel domains by initializing the system on DSQA, all the way to FSC, xSciQA and xDS, using the toolset derived from the HLE benchmark. HLE is selected as the foundational domain due to its substantial scale (2,500 queries) and extensive disciplinary coverage. Table~\ref{tab:knowledge_transfer} shows consistent performance across domains, with xSciQA scores improving from 76.5 to 80.2. Crucially, new tool synthesis drops by 55\% on FSC and 100\% on xSciQA and xDS, demonstrating effective transfer with minimal marginal cost.

\begin{table}[t]
    \centering
    \caption{Performance comparison and tool synthesis statistics across different initialization strategies. ZS stands for a zero start setting, and WS denotes warm start, which leverages the HLE-evolved toolset as initialization.}\label{tab:knowledge_transfer}
    \resizebox{\linewidth}{!}{\begin{tabular}{c|c|c|c}
        \midrule
        \textbf{Dataset} & \textbf{Strategy} & \textbf{Accuracy} & \textbf{\# New tools} \\
        \midrule
        HLE & ZS & 48.0 & 97 \\
        \midrule
        \multirow{2}{*}{DeepSearchQA} & ZS & 73.5 & 34 \\
        & WS & \phantom{$_{\uparrow 1.1}$}74.6$_{\color{red}\uparrow 1.1}$ & \phantom{$_{\downarrow 32\%}$}23$_{\color{green}\downarrow 32\%}$ \\
        \midrule
        \multirow{2}{*}{FinSearchComp} & ZS & 65.0 & 18 \\
        & WS & \phantom{$_{\uparrow 0.4}$}65.4$_{\color{red}\uparrow 0.4}$ & \phantom{$_{\downarrow 55\%}$}8$_{\color{green}\downarrow 55\%}$ \\
        \midrule
        \multirow{2}{*}{xbench-ScienceQA} & ZS & 76.5 & 13 \\
        & WS & \phantom{$_{\uparrow 3.7}$}80.2$_{\color{red}\uparrow 3.7}$ & \phantom{$_{\downarrow 100\%}$}0$_{\color{green}\downarrow 100\%}$ \\
        \midrule
        \multirow{2}{*}{xbench-DeepSearch} & ZS & 59.7 & 16 \\
        & WS & \phantom{$_{\uparrow 0.9}$}60.6$_{\color{red}\uparrow 0.9}$ & \phantom{$_{\downarrow 100\%}$}0$_{\color{green}\downarrow 100\%}$ \\
        \midrule
    \end{tabular}}
    \vspace*{-0.8cm}
\end{table}

\begin{figure}[t]
    \centering
    \includegraphics[width=\linewidth]{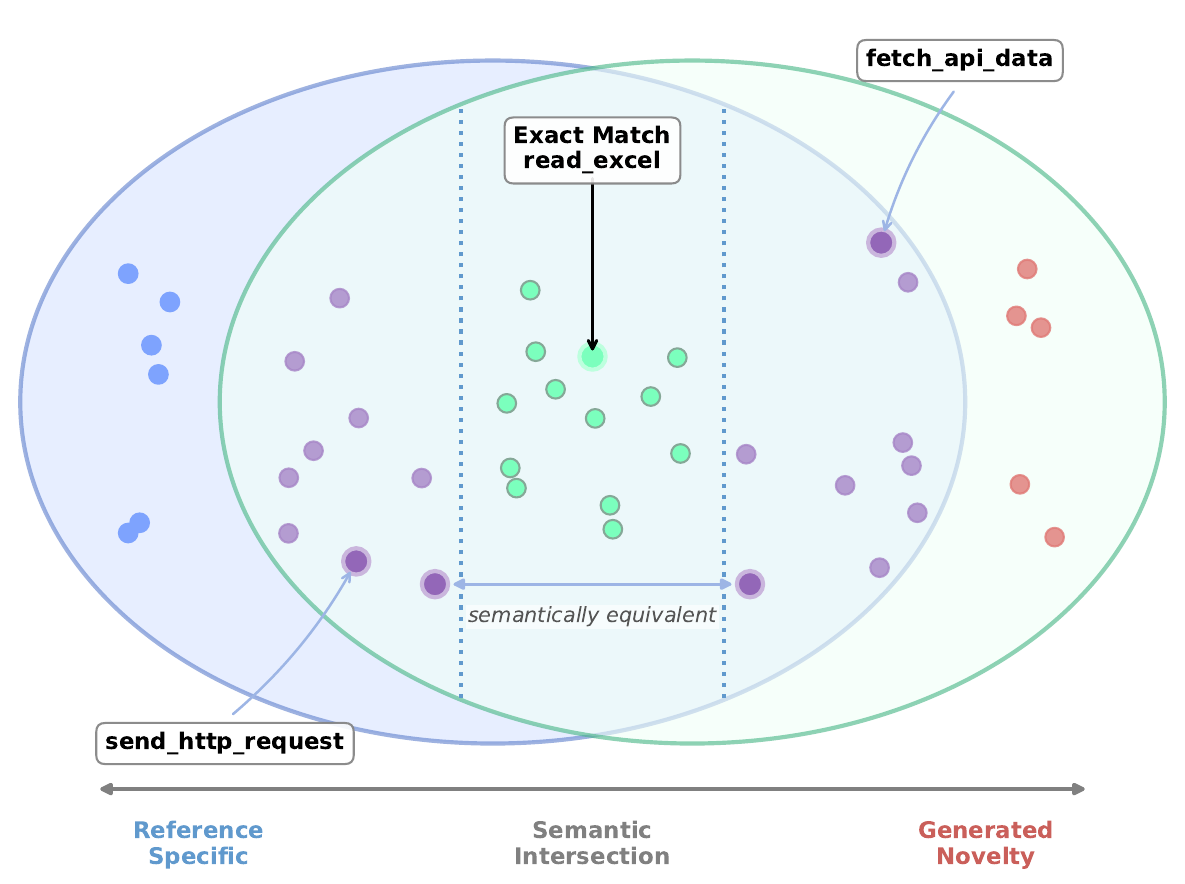}
    \vspace*{-0.5cm}
    \caption{Venn diagram comparing unique zero-start tools ($\mathcal{T}_{\text{DSQA}} \setminus \mathcal{T}_{\text{HLE}}$, left) and incremental warm-start tools ($\mathcal{T}_{\text{HLE}\rightarrow \text{DSQA}} \setminus \mathcal{T}_{\text{HLE}}$, right) on DeepSearchQA. Points represent semantically clustered tools. The intersection (71\%) denotes functional overlap, with dashed lines marking exact matches, while the difference set accounts for 29\%.}\label{fig:main_results_tool_diff_hle_dsqa}
    \vspace*{-0.3cm}
\end{figure}

\begin{figure*}[t]
    \centering
    \includegraphics[width=\linewidth]{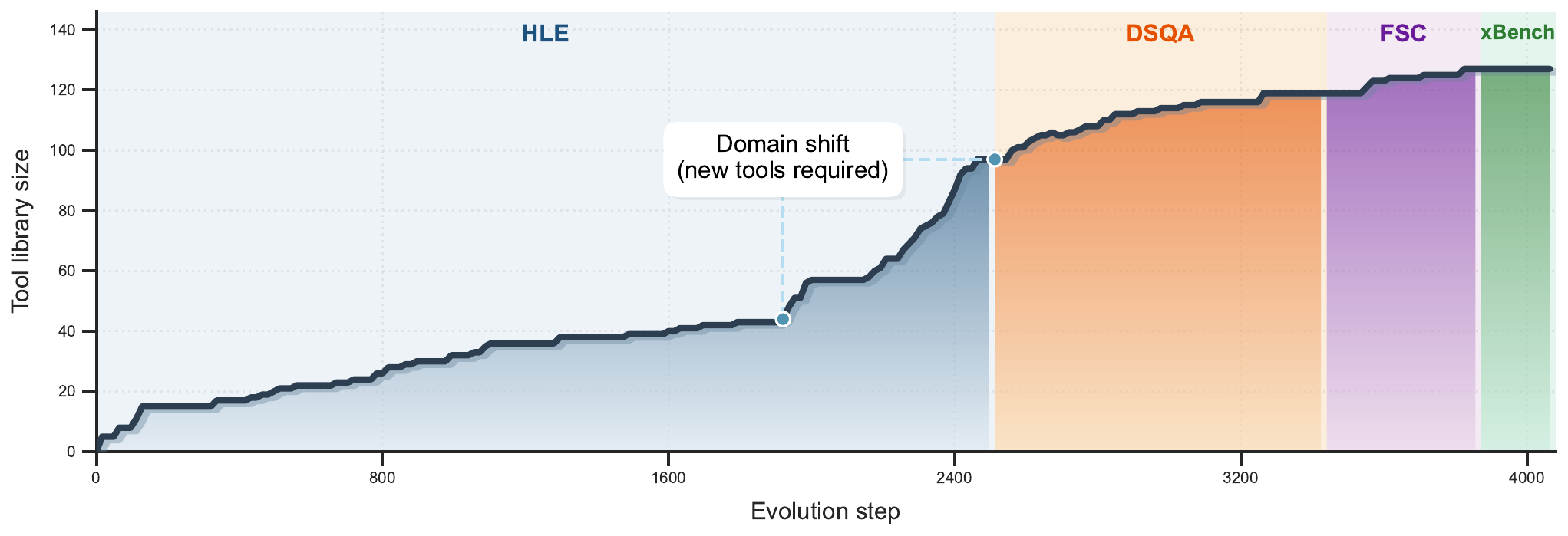}
    \vspace*{-0.5cm}
    \caption{Evolution of the tool library size relative to the cumulative number of processed queries. The experimental sequence follows the curriculum HLE $\rightarrow$ DeepSearchQA $\rightarrow$ FinSearchComp $\rightarrow$ xbench-ScienceQA $\rightarrow$ xbench-DeepSearch, highlighting the convergence of tool synthesis.}\label{fig:main_results_tool_num}
    \vspace*{-0.3cm}
\end{figure*}

Figure~\ref{fig:main_results_tool_num} illustrates the evolution of the tool library size relative to the cumulative number of processed queries. The overall trajectory demonstrates \textbf{a global trend toward convergence}. A localized surge in tool synthesis during the late phase of HLE (queries 2,000--2,400) is driven by intra-benchmark domain shifts: as query semantics transition from predominantly mathematical to social sciences and other disciplines, a targeted expansion of the toolset is necessitated. Notably, despite this semantic heterogeneity, \textbf{only 97 tools were generated across the entire 2,500-HLE query corpus, evidencing the effective consolidation of acquired knowledge}. Subsequently, tool growth trajectories across secondary datasets exhibit near-zero slopes, indicating that HLE-evolved capabilities facilitate \textbf{robust adaptation to novel domains with negligible marginal synthesis}.

To substantiate tools as vehicles for knowledge crystallization and domain adaptation, Figure~\ref{fig:main_results_tool_diff_hle_dsqa} visualizes the intersection between zero-start and warm-start toolsets. The recurrence of exact matches (e.g., \texttt{read\_excel}) validates the capacity to \textbf{deterministically recover essential utilities}. Furthermore, semantically aligned variants (e.g., \texttt{seed\_http\_request} vs. \texttt{fetch\_api\_data}, see Appendix~\ref{app:case_study_tool_similarity}) highlight adaptive functional equivalence. This convergence confirms the agent gravitates toward an \textbf{invariant core of capabilities independent of initialization}.

\begin{figure}[t]
    \centering
    \includegraphics[width=\linewidth]{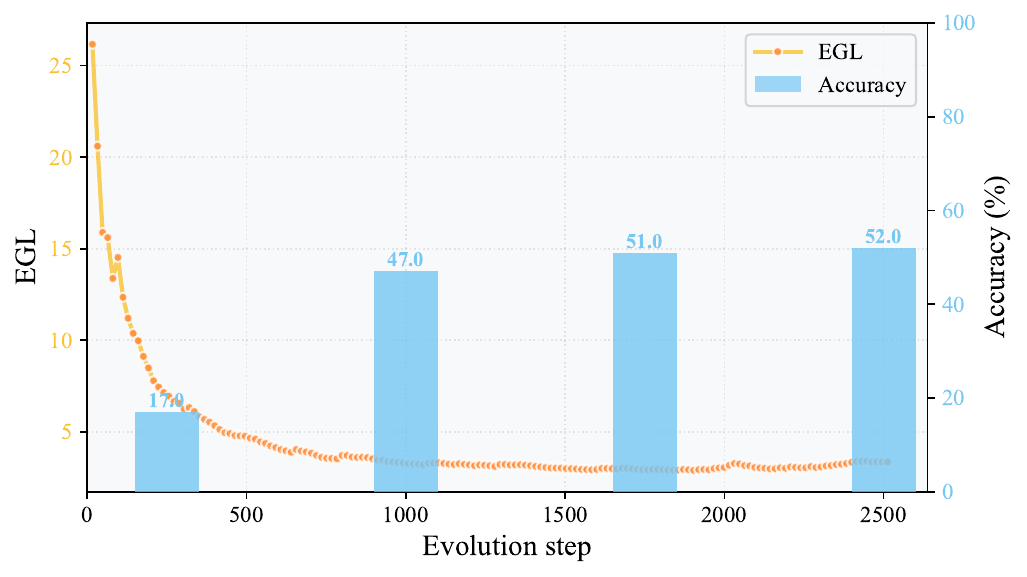}
    \vspace*{-0.5cm}
    \caption{EGL dynamics on HLE and accuracy on selected datasets vs. evolution step. The orange curve shows the EGL trend (left axis, scaled by 1000). Blue bars indicate the accuracy (right axis) of agents using HLE toolsets frozen at 10\%, 40\%, 70\%, and 100\% of evolution.}\label{fig:hle_evolution_convergence}
    \vspace*{-0.2cm}
\end{figure}

\subsection{Evolutionary generality loss}\label{sec:experiments_evolutionary_generality_loss}
To quantify knowledge acquisition, we introduce \textit{Evolutionary Generality Loss} (EGL), a real-time convergence indicator analogous to training loss. EGL is defined as the cumulative ratio of synthesized tools to invocations:
\vspace*{-0.2cm}
\begin{equation}
\text{EGL}(t) \triangleq \mathcal{L}_t = \frac{C_t}{U_t} = \frac{\sum_{i=1}^{t} c_i}{\sum_{i=1}^{t} u_i}
\end{equation}
where $c_i$ and $u_i$ denote tool synthesis and invocations at step $i$. EGL is inversely correlated with repository generality. As shown in Figure~\ref{fig:hle_evolution_convergence} (HLE dataset), initial high values reflect rapid accumulation, which stabilizes asymptotically after $\sim$1,000 queries (40\% of the stream).

\begin{figure*}[t]
    \centering
    \includegraphics[width=\linewidth]{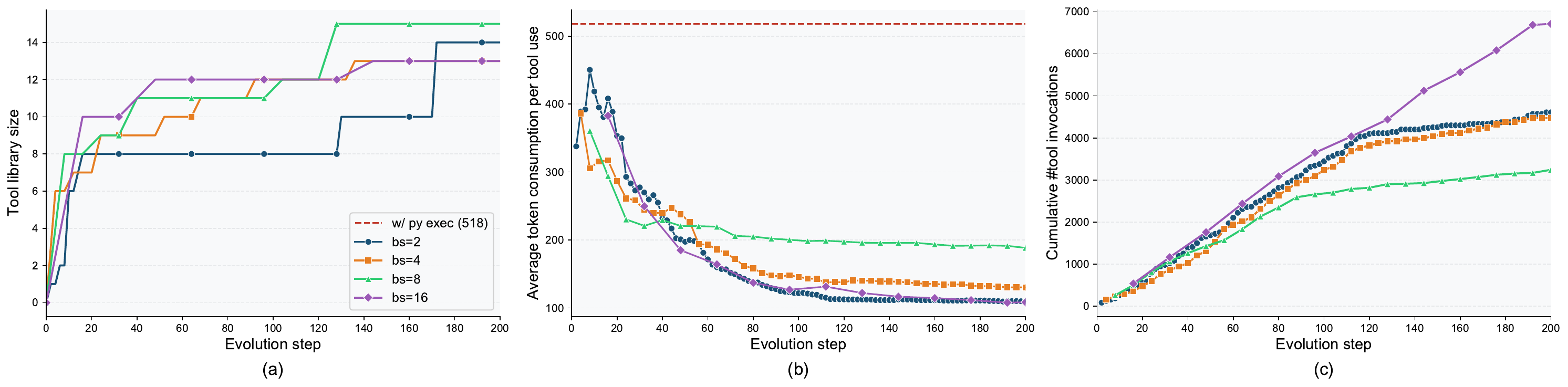}
    \vspace*{-0.5cm}
    \caption{Evolutionary dynamics across varying batch sizes. (a) Cumulative tool synthesis relative. (b) Mean token consumption per invocation. (c) Aggregate count of tool invocations.}\label{fig:ablation_batch_size_tool}
    \vspace*{-0.3cm}
\end{figure*}

Appendix~\ref{app:proof} provides the complete formal proofs. We first derive the sufficient condition for the monotonic decrease of EGL:
\begin{theorem}[Descent Condition]\label{thm:egl_decrease}
If the marginal synthesis ratio is strictly lower than the cumulative ratio (i.e., $\frac{c_{t+1}}{u_{t+1}} < \frac{C_t}{U_t}$), the EGL metric decreases ($\mathcal{L}_{t+1} < \mathcal{L}_t$).
\end{theorem}
Thus, EGL declines whenever the marginal creation rate falls below the historical average. Empirically (Figure~\ref{fig:hle_evolution_convergence}), the trajectory is consistently downward, with transient spikes reflecting domain-specific bursts.

Second, the convergence rate:
\begin{theorem}[Convergence Rate]\label{thm:egl_convergence}
If cumulative tools stabilize ($C_t \to C_\text{max}$) and invocations grow linearly ($U_t \propto t$), EGL converges to zero ($\mathcal{L}_t \to 0$) at a rate of $\mathcal{O}(1/t)$.
\end{theorem}
Figures~\ref{fig:main_results_tool_num} and~\ref{fig:ablation_batch_size_tool}(c) empirically validate the stabilization of $C_t$ and linear growth of $U_t$, respectively.

To further validate convergence, we evaluated toolset snapshots (frozen at 10\%, 40\%, 70\%, 100\% of the timeline) on 200 stratified queries with further tool synthesis disabled. Figure~\ref{fig:hle_evolution_convergence} bar chart reveals a sharp performance inflection between 10\%--40\%, followed by diminishing returns from 40\% to 100\%. This saturation confirms the system successfully converges to a high-generality state.

\begin{table}[t]
    \centering
    \definecolor{lightblue}{HTML}{E6F3FF}
    \definecolor{lightorange}{HTML}{FFF0E6}
    \caption{Ablation results on 200 queries across varying batch sizes (BS). Metrics include Accuracy (Acc, \%), total inference time in minutes (Time), Tool Success Rate (TSR, \%), Average Tokens per Tool (ATT), and cost per query (Cost, \$). The condition ``$\infty$'' denotes a Python-interpreter baseline processing queries independently without tool accumulation.}\label{tab:ablation_tool_evolution}
    {\begin{tabular}{c|c|c|c|c|c}
        \midrule
        \textbf{BS} & \textbf{Acc} & \textbf{Time} & \textbf{TSR} & \textbf{ATT} &\textbf{Cost} \\
        \midrule
        \rowcolor{lightblue} 2 & 48.0 & $\phantom{{}_{\div 1}}2,006_{\div 1}$ & 99.6 & 109.9 & 0.59 \\
        \rowcolor{lightblue} 4 & 50.0 & $\phantom{{}_{\div 2.1}}965_{\div 2.1}$ & 99.9 & 130.3 & 0.74 \\
        \rowcolor{lightblue} 8 & 47.5 & $\phantom{{}_{\div 6.2}}326_{\div 6.2}$ & 99.6 & 188.4 & 0.71 \\
        \rowcolor{lightblue} 16 & 51.5 & $\phantom{{}_{\div 9.1}}220_{\div 9.1}$ & 99.1 & 108.4 & 0.65 \\
        \midrule
        \rowcolor{lightorange} $\infty$ & 40.0 & $\phantom{{}_{\div 48.9}}41_{\div 48.9}$ & 81.8 & 518.2 & 0.71 \\
        \midrule
    \end{tabular}}
    \vspace*{-.6cm}
\end{table}

\subsection{Ablation study}
To investigate the efficacy of tool evolution, we conduct ablation studies using the 200-query EGL subset.

\textbf{Mechanisms of tool evolution efficacy.} We benchmark our approach against a Python-interpreter with the same baseline (GPT-5), representing the asymptotic limit of an infinite batch size ($B \to \infty$) where queries are processed independently without tool accumulation. As detailed in Table~\ref{tab:ablation_tool_evolution}, this baseline demonstrates markedly inferior performance (40.0) compared to our method (47.5--51.5). Analysis suggests this performance gap stems from the baseline's lower execution success rate (81.8\% vs. $>$99\%) and significantly higher token consumption (518 tokens per invocation). We posit that the resulting verbosity and frequent execution failures induce \textbf{contextual contamination}, which subsequently degrades reasoning capabilities (see Appendix~\ref{app:case_study}).

\begin{table}[t]
    \centering
    \caption{Performance comparison and tool library size of different backend models on DeepSearchQA and FinSearchComp.} \label{tab:ablation_backend}
    \resizebox{\linewidth}{!}{\begin{tabular}{c|cc|cc}
        \midrule
        \multirow{2}{*}{\textbf{Backend}} & \multicolumn{2}{c|}{\textbf{DeepSearchQA}} & \multicolumn{2}{c}{\textbf{FinSearchComp}} \\
        & \textbf{Accuracy} & \textbf{\# Tools} & \textbf{Accuracy} & \textbf{\# Tools} \\
        \midrule
        Gemini 3 Pro & 73.5 & 16 & 65.0 & 18 \\
        GPT-5 & 70.6 & 75 & 65.2 & 25 \\
        GPT-5-mini & 42.5 & 47 & 45.5 & 13 \\
        \midrule
    \end{tabular}}
    \vspace*{-.6cm}
\end{table}

\textbf{Impact of batch size on parallel evolution.} We investigate the influence of batch size $B$ on evolutionary dynamics, evaluating configurations where $B \in \{2, 4, 8, 16\}$ against a Python interpreter baseline ($B \to \infty$). As detailed in Table~\ref{tab:ablation_tool_evolution}, all batch-evolved agents surpass the baseline, reinforcing the efficacy of tool accumulation. Furthermore, the results indicate that increasing $B$ proportionally reduces processing latency and marginally enhances accuracy without incurring substantial overhead, thereby validating the \textit{Parallel Batch Evolution} strategy.

The impact of batch size on evolutionary trajectories is further analyzed in Figure~\ref{fig:ablation_batch_size_tool}(a). While larger batches accelerate initial tool crystallization, terminal library sizes remain comparable, suggesting the system \textbf{converges toward a consistent capability equilibrium} independent of $B$. Regarding computational efficiency, Figure~\ref{fig:ablation_batch_size_tool}(b) demonstrates that token expenditures consistently remain below the baseline, decaying toward an asymptotic minimum as the system transitions from costly \textit{ab initio} tool generation to efficient reuse. A minor anomaly observed for $B=8$ in Table~\ref{tab:ablation_tool_evolution}—characterized by higher token consumption—is attributed to maximal tool generation and invocation counts (Figure~\ref{fig:ablation_batch_size_tool}(a,c)) resulting from LLM stochasticity. Future research will explore methods to align convergence states across varying batch sizes.

\textbf{Influence of backend model selection.} We evaluate the impact of the underlying foundation model on system performance across the DSQA and FinSearchComp datasets (see Table~\ref{tab:ablation_backend}). The results indicate that when instantiated with \textit{GPT-5}, our agent surpasses the majority of established baselines. Notably, the lightweight \textit{GPT-5-mini} variant achieves competitive performance, demonstrating significant utility despite its reduced parameter scale. These findings underscore the model-agnostic nature and generalization capabilities of the proposed framework, confirming its robustness across diverse backend architectures and scales.

\section{Related work}

\textbf{Self-evolving agents}. To transcend the limitations of static designs, general-purpose agent systems must adapt to open-ended, interactive environments in real-time \citep{gao2025survey}. Self-evolving agents achieve this by autonomously refining internal components, categorized into workflow optimization, model parameter updates, context management, and tool synthesis. Specifically, MAS-GPT \citep{ye2025masgpt} and AFlow \citep{zhang2025aflow} generate query-specific workflows, while TT-SI \citep{acikgoz2025self} and SCA \citep{zhou2025selfchallenging} update parameters by generating challenging training problems. Furthermore, ELL \citep{cai2025building} and \citet{zhang2025agent} evolve by accumulating environmental interaction experience as context. Regarding tool synthesis, LIVE-SWE-agent \citep{xia2025live} and STELLA \citep{jin2025stella} enable on-the-fly tool creation for software engineering and biomedical research, respectively. Nevertheless, most self-evolving methods necessitate explicit training. Crucially, even approaches closest to our work—those autonomously creating tools at test-time—remain confined to specific domains and lack tool reuse mechanisms, hindering their general applicability.

\textbf{Tool evolution agents.} The paradigm of autonomous tool synthesis has garnered significant attention in recent literature. Pioneering systems such as Voyager~\citep{wang2023voyager}, STELLA~\citep{jin2025stella}, and LIVE-SWE-agent~\citep{xia2025live} empower agents to generate executable tools for embodied control or software engineering tasks. Notably, analyses of the tool embedding space in LIVE-SWE-agent~\citep{xia2025live} reveal distinct clustering patterns among functional equivalents; this observation directly informed the design of our \textit{absorbing mechanism}, which utilizes clustering to consolidate redundant utilities. Regarding tool reuse, Alita-G~\citep{qiu2025alitag} facilitates the persistence of successful tools, though its optimization trajectory is predicated on ground-truth supervision. In the context of inference-time adaptation, \citet{lu2026beyond} proposed a test-time tool evolution framework governed by a cost-sensitive objective function, yet its application remains narrowly scoped to scientific reasoning. Collectively, these approaches prioritize the optimization of existing assets or domain-specific synthesis, distinguishing them from our framework's focus on \textit{self-evolution in open-ended environments}.

\textbf{Agent systems for open-ended tasks}. Recent research has sought to construct general-purpose agents capable of cross-domain operation. DeepAgent \citep{li2025deepagent} utilizes a scalable toolset to address diverse tasks but lacks the capacity for autonomous environmental adaptation, leading to performance limitations. Agent KB \citep{tang2025agent} attempts to solve cross-domain problems by summarizing execution experiences from other agents; however, its performance is constrained by a deficiency in necessary tools. In contrast, our approach achieves generality by continuously creating, reusing, and optimizing tools in-situ, effectively overcoming the limitations of static toolsets and experience-based adaptation.

\section{Discussion and future work}
While this study establishes the efficacy of in-situ self-evolution through tool synthesis, several avenues for future research merit rigorous inquiry to fully realize the potential of autonomous agents.

\textbf{Paradigm parallels: towards system-level pre-training for agentic systems.} 
The seamless transition and performance consistency observed between our \textit{zero-start} and \textit{warm-start} settings offer more than just empirical validation; they suggest a fundamental paradigm shift. The clear convergence curves of the tool library indicate that ``task-solving capability'' is not merely a collection of ad-hoc heuristics, but a generalizable pattern that can be learned and distilled. 

This implies that the field is approaching a ``pre-training and post-training'' era for agentic systems, mirroring the trajectory of LLMs. We envision a future where multi-agent systems undergo \textit{system-level pre-training} on massive, broad-spectrum task datasets. This process would allow agents to distill a ``foundation toolset''—a set of converged, generalizable primitives—before deployment. Consequently, such pre-trained agents would possess intrinsic generalization capabilities, enabling them to tackle novel downstream tasks largely through the composition of existing reliable tools, thereby minimizing or even eliminating the need for expensive test-time evolution. Formalizing the methodologies for this ``agentic pre-training'' is a critical priority for future research.

\textbf{Co-evolution of memory and workflow.} Currently, our framework validates the self-evolving paradigm primarily through tool generation. However, tools constitute a necessary but insufficient condition for scenarios demanding high personalization or complex process management. For instance, personalized assistants require persistent, evolving memory structures to align with user preferences, while intricate tasks (e.g., deep research) necessitate structured workflow evolution—such as generating bespoke planning protocols for multi-agent coordination. A critical direction is extending the evolutionary mechanism to encompass the co-evolution of memory architectures and workflow policies, enabling the system to adapt its internal state and execution logic alongside its functional capabilities.

\textbf{Evolutionary stability and regularization.} The stability of tool evolution warrants further investigation. Due to the inherent stochasticity of LLM generation, toolsets can exhibit variance across experimental runs, as evidenced by our batch size ablation. Ensuring the consistent convergence of the tool library is vital for system reliability. Future work will focus on developing regularization strategies to guarantee the determinism of the evolutionary process in open-ended environments.

\textbf{Optimization of parallel batch evolution.} Finally, the nuances of the batch evolution strategy present rich opportunities for optimization:
\begin{itemize}
    \vspace*{-0.3cm}
    \item \textit{Curriculum learning effects:} As shown in Figure~\ref{fig:main_results_tool_num}, the sequence of incoming queries significantly influences the convergence trajectory; delaying queries that require foundational primitives can impede system maturation. Investigating optimal query ordering is a key direction.
    \vspace*{-0.15cm}
    \item \textit{Intra-batch diversity trade-offs:} Intra-batch diversity presents a complex dynamic. Low diversity allows the \textit{absorbing mechanism} to function as a form of \textit{Best-of-N test-time scaling}, leveraging redundancy to select the optimal implementation—a contrast to traditional gradient training where high batch diversity is preferred. Balancing this quality-assurance benefit against the efficiency of capability evolution is crucial.
    \vspace*{-0.15cm}
    \item \textit{Adaptive scheduling:} Dynamic batch sizing offers a promising avenue for optimization. Smaller batch sizes in early stages could foster the robust consolidation of general knowledge via redundancy, while larger batches in the post-convergence phase could maximize throughput for corner cases. Developing autonomous agents capable of dynamic batch scheduling based on convergence signals remains a worthy research direction.
\end{itemize}

\section{Conclusion}
In this work, we presented the In-Situ Self-Evolving framework, enabling LLM-based agents to autonomously adapt and evolve within open-ended environments. Through the Yunjue Agent and the proposed Parallel Batch Evolution strategy, we demonstrated that treating tools as dynamic vehicles for knowledge crystallization allows for robust zero-start learning and the emergence of generalized, transferable capabilities. Our empirical results validate that this approach achieves state-of-the-art performance across heterogeneous benchmarks. Furthermore, warm-start experiments confirm that the accumulated generalized knowledge can be seamlessly transferred to novel domains, enabling the agent to adapt to shifting environments. We release our codebase, evaluation and system traces to the open-source community to facilitate further research. Broadly, future work will aim to unify tool evolution with memory and workflow adaptation, advancing toward more autonomous and general-purpose agentic systems.

\bibliographystyle{icml2025}
\bibliography{custom}

\onecolumn
\appendix
\raggedbottom
\section{Mathematical proof}\label{app:proof}
This section details the formal proofs for Theorems~\ref{thm:egl_decrease} and~\ref{thm:egl_convergence}, originally presented in Section~\ref{sec:experiments_evolutionary_generality_loss}.

\subsection{Proof of theorem~\ref{thm:egl_decrease}}
We begin by establishing the sufficient condition for the monotonic decrease of the EGL metric at step $t+1$ (i.e., $\mathcal{L}_{t+1} < \mathcal{L}_t$). By definition:
\begin{equation}
\mathcal{L}_{t+1} = \frac{C_t + c_{t+1}}{U_t + u_{t+1}} < \frac{C_t}{U_t} = \mathcal{L}_t
\end{equation}

Since the total invocation count $U_t$ is strictly positive, we cross-multiply without reversing the inequality:
\begin{equation}
(C_t + c_{t+1}) \cdot U_t < C_t \cdot (U_t + u_{t+1})
\end{equation}

Expanding the terms yields:
\begin{equation}
C_t U_t + c_{t+1} U_t < C_t U_t + C_t u_{t+1}
\end{equation}

Subtracting the common term $C_t U_t$ and rearranging results in the \emph{Descent Condition}:
\begin{equation}
\frac{c_{t+1}}{u_{t+1}} < \frac{C_t}{U_t}
\end{equation}

This derivation demonstrates that $\mathcal{L}_t$ decreases if and only if the marginal ratio of synthesis to invocation at the current step ($c_{t+1}/u_{t+1}$) is strictly lower than the historical cumulative ratio ($C_t/U_t$). During early evolutionary stages, extensive exploration ($c_{t+1} > 0$) may induce fluctuations. However, as the agent transitions to an exploitation phase ($c_{t+1} \to 0$ while $u_{t+1} \ge 1$), the marginal ratio vanishes, guaranteeing satisfaction of the inequality and ensuring the strict decrease of the metric.

\subsection{Proof of theorem~\ref{thm:egl_convergence}}

To establish the asymptotic dynamics of $\mathcal{L}_t$, we introduce the following assumptions governing the evolutionary process:

\begin{assumption}[Knowledge Boundedness]\label{asmp:knowledge_boundedness}
The task domain admits a finite set of optimal primitives sufficient to resolve all potential queries. As the agent evolves, the cumulative count of synthesized tools $C_t$ converges to a finite upper bound $C_\text{max}$. Consequently, marginal tool generation vanishes asymptotically:
\begin{equation}
\lim_{t \to \infty} C_t = C_\text{max} \implies \lim_{t \to \infty} c_t = 0
\end{equation}
\end{assumption}
\noindent We posit this as a mild condition. Empirically, as validated in Figure~\ref{fig:hle_evolution_convergence}, the marginal growth of synthesized tools diminishes toward zero once the agent consolidates sufficient generalized knowledge.

\begin{assumption}[Continuous Activity]\label{asmp:continuous_activity}
We assume every newly synthesized tool is immediately invoked at least once, ensuring $0 \le c_t \le u_t$ and $0 \le \frac{c_t}{u_t} \le 1$. Operationally, if a synthesized tool remains uninvoked, it is discarded post-processing to satisfy this constraint. Additionally, we assume every query necessitates at least one tool invocation ($u_t \ge 1$), with the invocation count fluctuating around a stationary mean $\mu \ge 1$. This implies that cumulative invocations $U_t$ grow linearly with respect to $t$:
\begin{equation}
U_t \approx \mu \cdot t \quad \text{as } t \to \infty
\end{equation}
\end{assumption}
\noindent Note that for the boundary condition $u_t = 0$, we define $c_t / u_t = 0$ to avoid singularity. Figure~\ref{fig:ablation_batch_size_tool}(c) provides empirical evidence supporting the linear growth of $U_t$ throughout the evolutionary steps.

Based on Assumptions~\ref{asmp:knowledge_boundedness} and~\ref{asmp:continuous_activity}, we analyze the limit of $\mathcal{L}_t$ as $t \to \infty$:
\begin{equation}
\lim_{t \to \infty} \mathcal{L}_t = \frac{\lim_{t \to \infty} C_t}{\lim_{t \to \infty} U_t} \approx \lim_{t \to \infty} \frac{C_\text{max}}{\mu \cdot t}
\end{equation}

Since $C_\text{max}$ and $\mu$ are positive constants:
\begin{equation}
\lim_{t \to \infty} \mathcal{L}_t = 0
\end{equation}

This confirms that EGL functions mathematically as a loss metric, converging to zero at a rate of $\mathcal{O}(1/t)$ as system stability is achieved.

\section{Algorithm}\label{app:algorithm}
\begin{algorithm}[h]
   \caption{Yunjue Agent Execution and Evolution Pipeline}
   \label{alg:ise_pipeline}
\begin{algorithmic}[1]
   \STATE {\bfseries Input:} Query Batch $\mathcal{Q}_t = \{q_{t,1}, \dots, q_{t,B}\}$, Global Tool Library $\mathcal{T}_{t-1}$
   \STATE {\bfseries Output:} Answers $\mathcal{A}_t$, Updated Library $\mathcal{T}_t$
   \STATE $\mathcal{A}_t \gets \emptyset$, $\mathcal{T}_{new} \gets \emptyset$
   \FOR{\textbf{each} query $q_{t,i} \in \mathcal{Q}_t$ \textbf{in parallel}}
       \STATE $\mathcal{R}_{miss} \gets \emptyset$, $\text{done} \gets \textbf{false}$
       \WHILE{\textbf{not} $\text{done}$}
           \STATE $\omega_{req} \gets \textbf{Manager}(q_{t,i}, \mathcal{T}_{t-1}, \mathcal{R}_{miss})$ \COMMENT{Analysis required tools}
           \IF{new tool required}
               \STATE $\mathcal{P}_{t,i} \gets \textbf{ToolDeveloper}(q_{t,i}, \omega_{req})$ \COMMENT{Synthesize new tools}
               \STATE $\mathcal{T}_{local} \gets \mathcal{T}_{t-1} \cup \mathcal{P}_{t,i}$
               \STATE $\mathcal{T}_{new} \gets \mathcal{T}_{new} \cup \mathcal{P}_{t,i}$
           \ELSE
               \STATE $\mathcal{T}_{local} \gets \mathcal{T}_{t-1}$
           \ENDIF
           \STATE $r_{t,i}, \mathcal{R}_{miss} \gets \textbf{Executor}(q_{t,i}, \mathcal{T}_{local})$ \COMMENT{Execute and update capability missing report}
           \IF{$\mathcal{R}_{miss} = \emptyset$}
               \STATE $\text{done} \gets \textbf{true}$
           \ENDIF
       \ENDWHILE
       \STATE $a_{t,i} \gets \textbf{Integrator}(q_{t,i}, r_{t,i})$ \COMMENT{Integrate final answer}
       \STATE $\mathcal{A}_t \gets \mathcal{A}_t \cup \{a_{t,i}\}$
   \ENDFOR
   \STATE $\mathcal{C}_{clusters} \gets \textbf{Aggregator}(\mathcal{T}_{t-1}, \mathcal{T}_{new})$ \COMMENT{Cluster and absorb analogous tools}
   \STATE $\mathcal{T}_t \gets \textbf{Merger}(\mathcal{C}_{clusters})$ \COMMENT{Merge and update global repository}
   \STATE \textbf{return} $\mathcal{A}_t$, $\mathcal{T}_t$
\end{algorithmic}
\end{algorithm}

\section{Experiment setup}\label{app:setup}

\begin{table}[h]
    \centering
    \caption{Statistics of the datasets.}\label{tab:dataset_stats}
    \begin{tabular}{l|c|c|c}
    \toprule
    \textbf{Dataset} & \textbf{Domain} & \textbf{Language} & \textbf{\# Queries} \\
    \midrule
    HLE & General reasoning & EN & 2,500 \\
    DeepSearchQA & General QA & EN & 900 \\
    FinSearchComp (T1 \& T2) & Financial search & EN \& CH & 391 \\
    xBench-ScienceQA & General science & CH & 100 \\
    xBench-DeepSearch & General research & CH & 100 \\
    \bottomrule
    \end{tabular}
\end{table}

\begin{figure}[t]
    \centering
    \includegraphics[width=.7\linewidth]{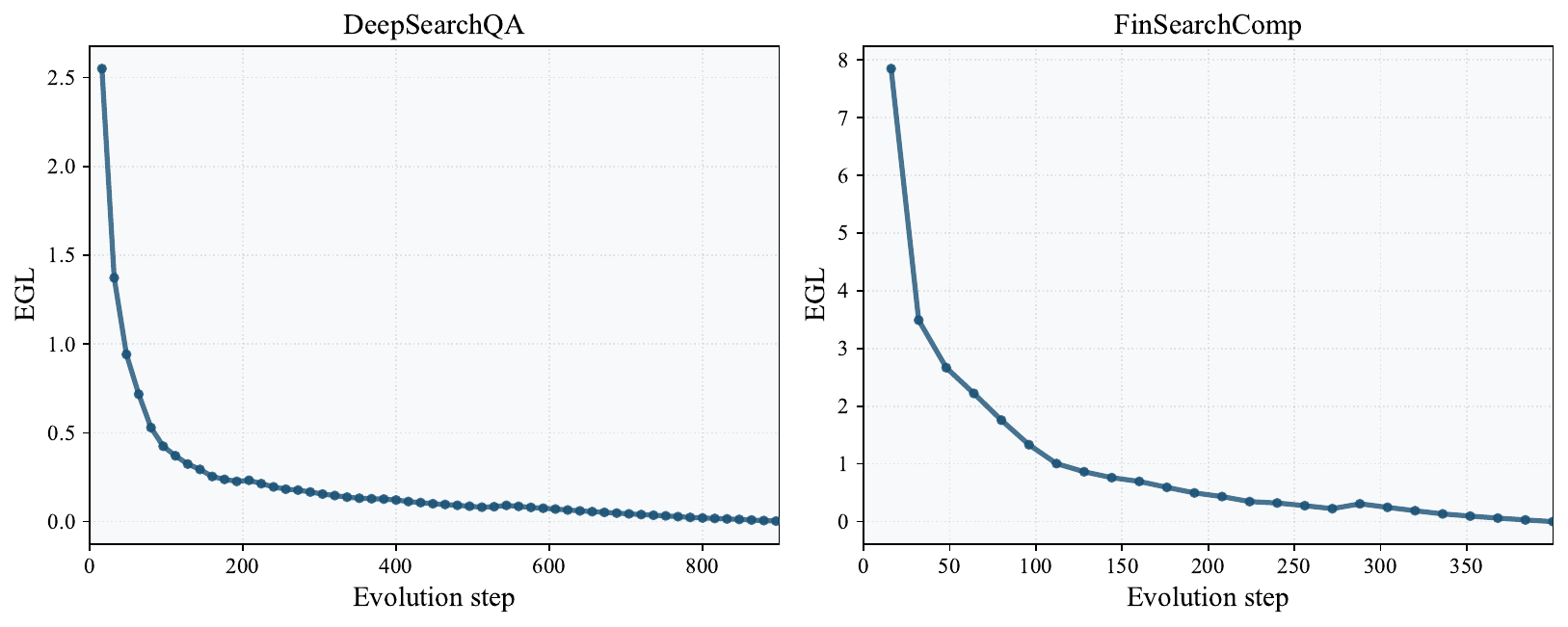}
    \vspace*{-0.3cm}
    \caption{EGL dynamics on the DSQA and FSC datasets.}\label{fig:egl_dsqa_fsc}
\end{figure}

Data statistics are summarized in Table~\ref{tab:dataset_stats}.

\textbf{Evaluation protocols.} We follow the standard evaluation protocols released with each dataset:
\begin{itemize}
    \item HLE: \href{https://github.com/centerforaisafety/hle/blob/main/hle_eval/run_judge_results.py}{Official evaluation script}
    \item DeepSearchQA: \href{https://www.kaggle.com/code/andrewmingwang/deepsearchqa-starter-code}{Official evaluation script}
    \item FinSearchComp: \href{https://huggingface.co/datasets/ByteSeedXpert/FinSearchComp}{Judging prompts}
    \item xBench: \href{https://github.com/xbench-ai/xbench-evals/blob/main/eval_grader.py}{Official evaluation script}
\end{itemize}

\textbf{Baselines sources.} We collect baseline results from their official reports or leaderboards. For HLE, results for Gemini Deep Research and Gemini 3 Pro are from~\cite{google2025deepresearch}; MiroThinker-v1.0-72B and Tongyi DeepResearch are from~\cite{team2025mirothinker}; GPT-5.2 Pro is from~\cite{openai2025gpt52}; Seed-1.8 is from~\cite{seed18modelcard}; Claude 4.5 Opus is from~\cite{claude45opus}; and Co-Sight is from~\cite{cosight}. DeepSearchQA results are from~\cite{deepsearchqa}. FinSearchComp results are from~\cite{seed18modelcard}. For xBench, results on ScienceQA and DeepSearch are retrieved from~\cite{xbenchscienceqa} and~\cite{xbenchdeepsearch}, respectively. Additionally, for xBench-DeepSearch, we include M-GRPO~\cite{mgrpo} and AWorld~\cite{aworld}.

\textbf{Implementation details.} To ensure experimental consistency, all nodes in our system are instantiated using the same \textit{backend LLM}: Gemini 3 Pro. The batch size is set to $B=16$ for all datasets. The \textit{Tool Developer} module is explicitly powered by Codex~\cite{openai2021codex} to facilitate robust code generation. We main a fixed temperature of 0.7 across all LLM invocations. To simplify the use of multimodal capabilities, we encapsulate image processing functionality into a dedicated tool, ensuring utilization of the same backend LLM. Our agent system is built based on LangGraph~\cite{langgraph}. All prompts in our system adopt the Markdown template format supported by Jinja~\cite{jinja}.

\section{More evaluation results}\label{app:results}
\begin{table}[t]
    \centering
    \caption{Experimental results across five benchmarks of the Yunjue Agent in the zero-start setting. ``Token'' denotes the average token consumption per query, ``Exec. Ratio'' indicates the proportion of tokens consumed by the Executor, and ``Cost'' denotes the monetary expense based on token usage.}\label{tab:main_results}
    \begin{tabular}{c|c|c|c|c}
        \midrule
        \textbf{Dataset} & \textbf{Accuracy} (\%)  & \textbf{Token} & \textbf{Exec. Ratio} (\%) & \textbf{Cost} (\$) \\
        \midrule
        HLE & $48.0_{\pm 0.2}$ & 421k & 95.1 & 1.48 \\
        DeepSearchQA & $73.5_{\pm 0.4}$ & 718k & 95.6 & 1.78 \\
        FinSearchComp & $65.0_{\pm 0.1}$ & 474k & 94.5 & 1.30 \\
        xbench-ScienceQA & $76.5_{\pm 0.6}$ & 207k & 94.2 & 0.73 \\
        xbench-DeepSearch & $59.7_{\pm 0.2}$ & 569k & 94.7 & 1.48 \\
        \midrule
    \end{tabular}
\end{table}

\textbf{Detailed results on five benchmarks.} Table~\ref{tab:main_results} summarizes the detailed performance of the Yunjue Agent in the zero-start setting across five datasets, where accuracy is calculated as the mean and standard deviation over three runs using the LLM-as-judge.

\begin{figure}[t]
    \centering
    \includegraphics[width=\linewidth]{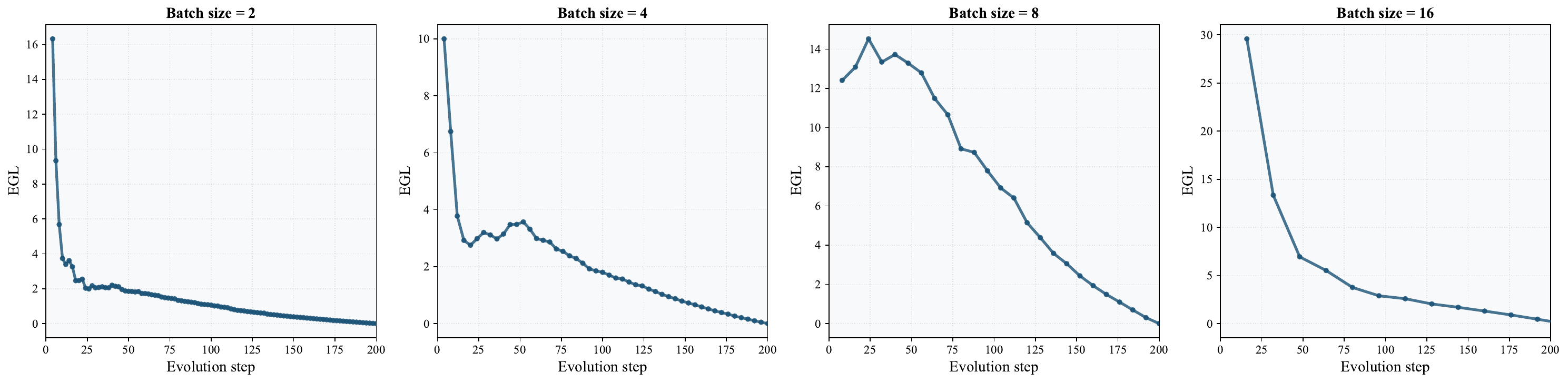}
    \vspace*{-0.3cm}
    \caption{EGL dynamics under different batch sizes.}\label{fig:egl_batch_size}
    \vspace*{-0.2cm}
\end{figure}

\textbf{More results on evolutionary generality loss (EGL).} Figure~\ref{fig:egl_dsqa_fsc} tracks the EGL trajectories across the DSQA and FSC benchmarks, while Figure~\ref{fig:egl_batch_size} dissects the impact of batch size on optimization behavior. Collectively, these visualizations validate the asymptotic stability of our approach across varying experimental conditions.

\textbf{Technical insights.} We observed that Gemini 3 Pro possesses strong reasoning capabilities and internal knowledge, often exhibiting a sense of ``confidence'', which is manifested by its tendency to rely on fewer tools to complete tasks. However, it frequently suffers from hallucinations and often exhibits weaker instruction-following capabilities, necessitating iterative prompt engineering. In contrast, GPT-5 and GPT-5-mini often appear more ``cautious'', typically requiring the invocation of more tools, planning, and reasoning steps to iteratively verify task results. This leads to increased tool creation and longer execution histories. This phenomenon is particularly pronounced in GPT-5 (as shown in Table~\ref{tab:ablation_backend}, the GPT-5 series clearly creates more tools). Nevertheless, the GPT-5 series demonstrates exceptional instruction-following abilities; adding constraints to their prompts generally yields behavior consistent with expectations.

\begin{figure}[t]
    \centering
    \includegraphics[width=\linewidth]{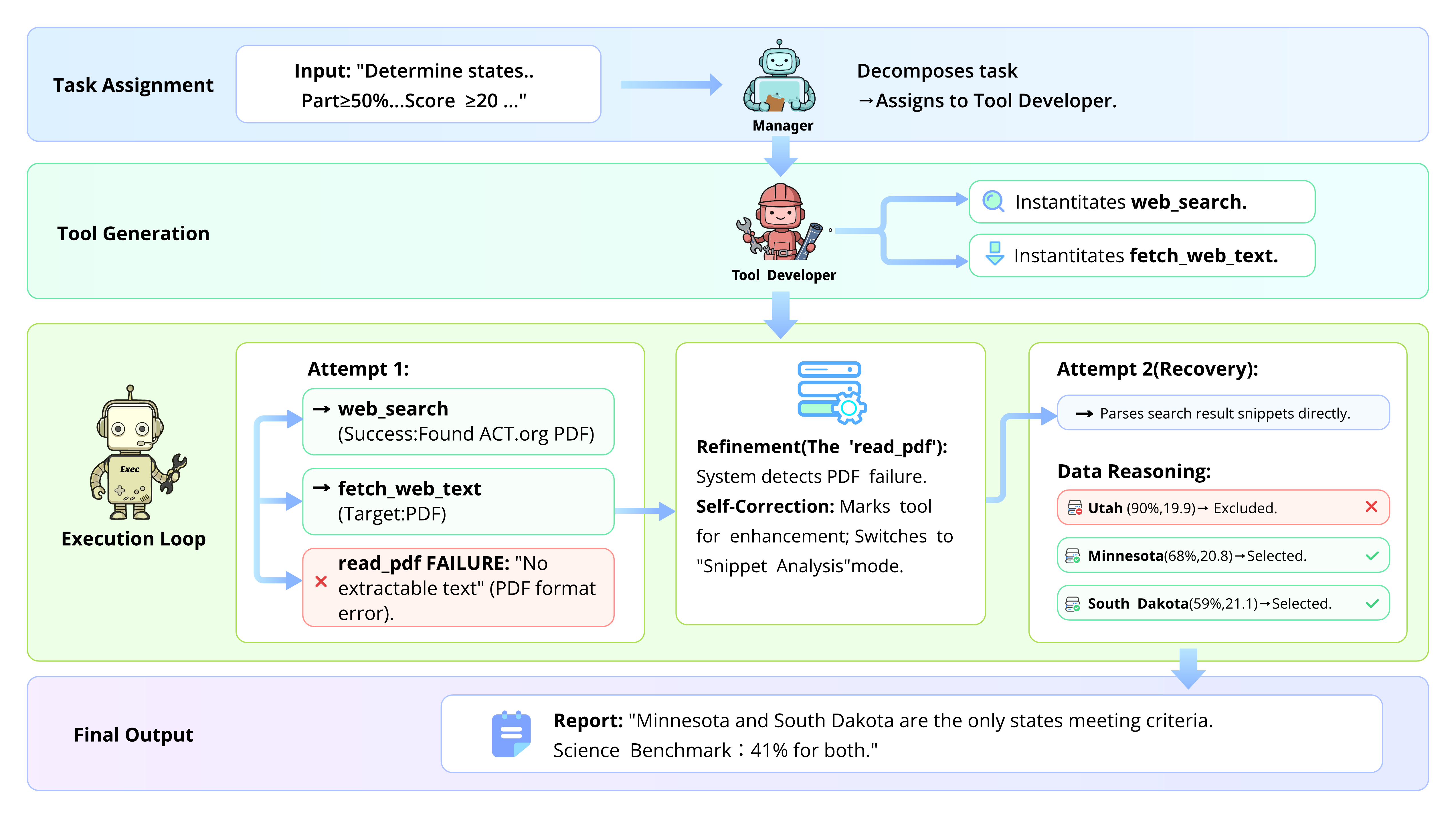}
    \vspace*{-0.8cm}
    \caption{Yunjue Agent execution pipeline. Upon receiving a query, the agent autonomously synthesizes, deploys, and iteratively refines tools to derive the final response, seamlessly integrating generation and execution.}\label{fig:case_study_tool_create_refine}
    \vspace*{-0.3cm}
\end{figure}

\begin{figure}[t]
    \centering
    \includegraphics[width=\linewidth]{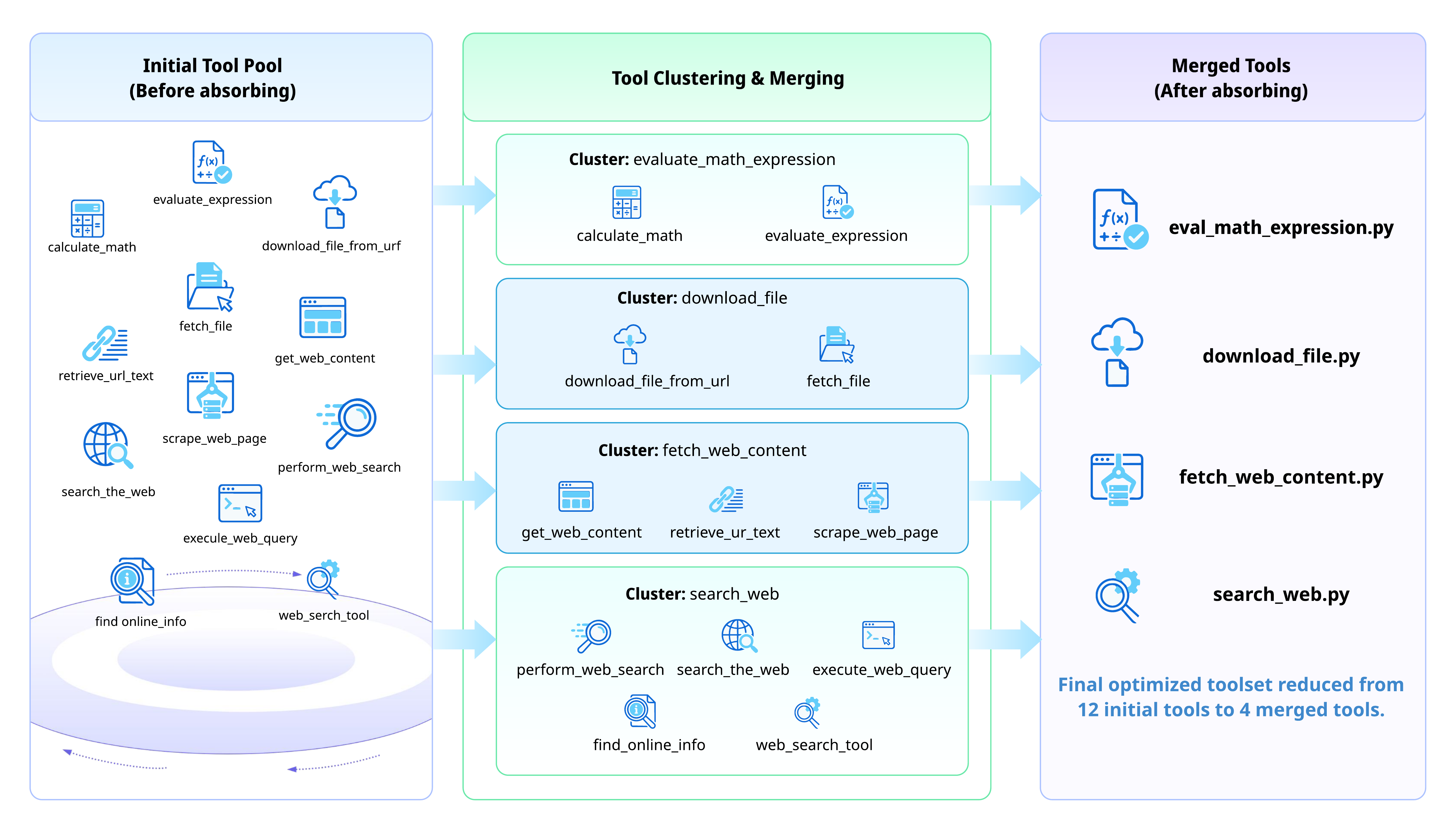}
    \vspace*{-0.7cm}
    \caption{Illustration of tool absorbing mechanism. Following batch execution, functionally analogous tools are identified via clustering and consolidated into a generalized, compact toolset, effectively pruning redundancy.}\label{fig:case_study_absorbing}
    \vspace*{-0.2cm}
\end{figure}

\begin{figure}[h!]
    \centering
    \includegraphics[width=\linewidth]{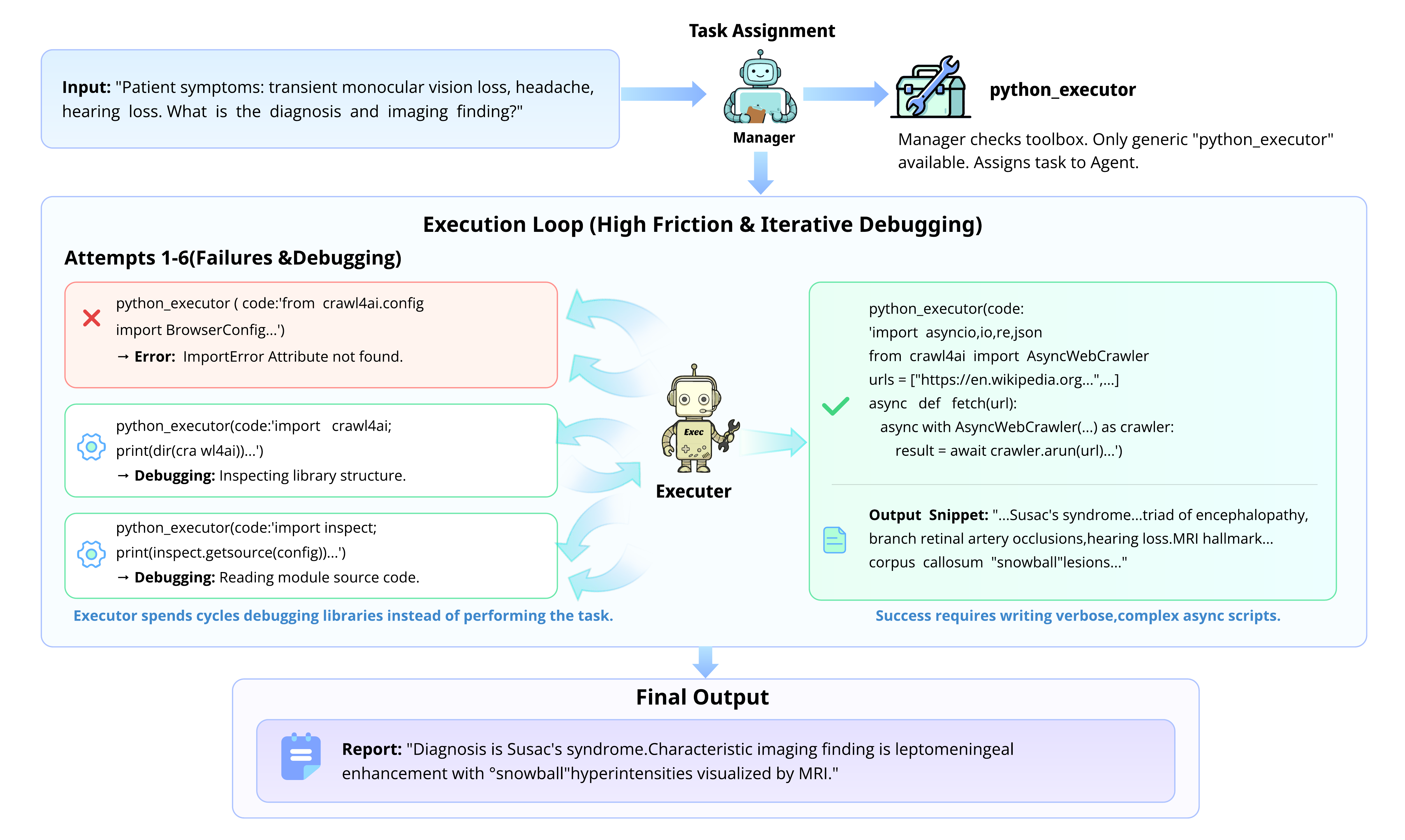}
    \vspace*{-0.7cm}
    \caption{Python-only baseline failure mode. Unlike our approach, the baseline indiscriminately accumulates raw execution traces—including erroneous attempts—within the context window, leading to severe contextual contamination and reasoning degradation.}\label{fig:case_study_py_exec}
    \vspace*{-0.2cm}
\end{figure}

\section{Case study}\label{app:case_study}

\textbf{Tool evolution.} Figure~\ref{fig:case_study_tool_create_refine} illustrates the end-to-end execution pipeline of the Yunjue Agent. When presented with a query, the \textit{Manager} delegates tool creation to the \textit{Tool Developer}. The agent effectively instantiates necessary tools and, crucially, demonstrates self-correction capabilities. When an initial tool fails (e.g., a PDF reader failure), the system detects the error, refines its strategy, and successfully derives the final answer, highlighting the seamless integration of tool generation, execution, and refinement.

\textbf{Knowledge consolidation via absorbing.} Figure~\ref{fig:case_study_absorbing} depicts the tool absorbing mechanism. Following batch execution, the initial tool pool often contains redundant or highly specific tools. The system employs clustering to identify functionally analogous tools and merges them into a compact, generalized toolset (e.g., consolidating multiple web scrapers into a single robust tool). This process effectively prunes redundancy, ensuring the long-term efficiency and generalizability of the tool library.

\textbf{Comparison with Python-only baseline.} Figure~\ref{fig:case_study_py_exec} contrasts our approach with a typical Python-only baseline failure mode. In complex tasks like medical diagnosis, the baseline often struggles with library dependencies and engages in high-friction debugging loops. Unlike our approach, the baseline accumulates raw, erroneous execution traces within its context window, leading to severe context contamination and reasoning degradation.

\textbf{Tool similarity and clustering}\label{app:case_study_tool_similarity}
To verify the semantic alignment of tools between independent evolutionary runs (as visualized in Figure~\ref{fig:main_results_tool_diff_hle_dsqa}), we present a direct code comparison. We contrast \texttt{send\_http\_request}, under a zero-start setting, with \texttt{fetch\_api\_data}, under a warm-start setting, evolved for DeepSearchQA. Despite distinct origins, both converge on core HTTP functionality, necessitating similar dependencies and logic structures.

\definecolor{toolblue}{HTML}{75c7f4}

\begin{tcblisting}{
    listing only,
    breakable,
    colback=white,
    colframe=toolblue,
    coltitle=white,
    title=\textbf{Tool: send\_http\_request},
    fonttitle=\bfseries\large,
    arc=2pt,
    boxrule=1.5pt,
    left=6pt, right=6pt, top=6pt, bottom=6pt,
    enhanced,
    listing options={
      breaklines=true,
      basicstyle=\ttfamily\scriptsize,
      columns=fullflexible,
      language=Python
    }
}
__TOOL_META__ = {
    "name": "send_http_request",
    "description": "Sends customizable HTTP GET or POST requests to a specified URL with optional form or JSON payloads, allowing the inclusion of custom headers and returning the response status code plus text-based body content or a saved-file notice for binary responses.",
    "dependencies": ["requests", "beautifulsoup4"],
}

try:
    response = session.request(method=input.method, url=input.url, **request_kwargs)
except requests.RequestException as exc:
    return OutputModel(status_code=-1, response_text=f"Request failed: {exc}")

content_type = response.headers.get("Content-Type", "").lower()
if content_type and _is_textual_content(content_type):
    body_text = response.text
elif not content_type and response.encoding:
    body_text = response.text
\end{tcblisting}
\begin{tcblisting}{
    listing only,
    breakable,
    colback=white,
    colframe=toolblue,
    coltitle=white,
    title=\textbf{Tool: fetch\_api\_data},
    fonttitle=\bfseries\large,
    arc=2pt,
    boxrule=1.5pt,
    left=6pt, right=6pt, top=6pt, bottom=6pt,
    enhanced,
    listing options={
      breaklines=true,
      basicstyle=\ttfamily\scriptsize,
      columns=fullflexible,
      language=Python
    }
}
__TOOL_META__ = {
    "name": "fetch_api_data",
    "description": "Performs an HTTP GET request to a specified API endpoint with optional query parameters and headers, then returns the raw response body (typically JSON) for downstream programmatic consumption.",
    "dependencies": ["httpx"],
}

with httpx.Client(timeout=timeout) as client:
    response = client.get(input.url, params=params, headers=headers)
    response.raise_for_status()
    content_type = response.headers.get("content-type", "").lower()
    if "application/json" in content_type:
        body = response.text
    elif "text/" in content_type:
        body = response.text
    else:
        body = response.text
        OutputModel(response_data=body)
\end{tcblisting}

\clearpage
\section{Prompts}\label{app:prompts}
\definecolor{customblue}{HTML}{75c7f4}

\begin{tcblisting}{
    listing only,
    breakable,
    colback=white,
    colframe=customblue,
    coltitle=white,
    title=\textbf{Manager},
    fonttitle=\bfseries\large,
    arc=2pt,
    boxrule=1.5pt,
    left=6pt, right=6pt, top=6pt, bottom=6pt,
    enhanced,
    listing options={
      breaklines=true,
      basicstyle=\ttfamily\scriptsize,
      columns=fullflexible,
      inputencoding=utf8,
      extendedchars=true
    }
}
You are a Task Orchestrator. Your mission is to analyze the task, determine the exact set of tools needed--selecting from available ones or defining new ones if absolutely necessary, and provide a strategic outline for how these tools can be used to complete the task.

# Core Principle
**Your absolute priority is to enable the executor to complete the `Task` through clear guidance and the right tools.** You must prioritize the combination of available, atomic tools. Only request new tools if this task can not be achieved using the available tools (even by chaining them). **NEVER create a composite tool that merely combines two or more available atomic tool capabilities.**

# Task Information
## Task
{{ user_query }}

{% if failure_report %}
## Failure Report For Previous Execution
{{ failure_report }}
{% endif %}

{% if additional_tool_requests %}
## Tool Request from Executor
{{ additional_tool_requests }}
{% endif %}

# Available Tools

The following tools are currently available:

{% for tool in available_tools %}
- **{{ tool.name }}**: {{ tool.description }}. The input args is {{ tool.input_schema }}
{% endfor %}

# Analysis Instructions

1. **Analyze the task requirements**: You must carefully think through which tools are needed to complete this task. {% if failure_report %} **You need to pay close attention to the content of suggestions and consider whether there are new tools that can help the executor complete the task.** 
{% endif %}

2. **Exhaustive available Tool Check (Priority #1)**:
  - **Ask yourself:** "Can I accomplish this task by using available tools?"
  - **Ask yourself:** "If multiple similar tools exist that can accomplish a specific objective, which tool is the best for this scenario based on their descriptions?"
    - **Example:** If you need "weather for Paris", and you have a tool for web searching, USE this one. DO NOT create `get_weather`.
  - **Tool name fidelity (MUST, case-sensitive)**:
    - You MUST treat tool names as **exact identifiers**. In your final JSON, every entry in `required_tool_names` MUST be copied **verbatim** (character-for-character) from the provided `Available Tools` list.
  - **Image Capabilities Priority:** If the task requires OCR, image understanding, or other image processing capabilities, **prioritize using the `image_text_query` tool**.
  - **Data Access Strategy:** Prefer **search/filter**, **metadata/summary inspection**, and **bounded previews / range reads** (with explicit limits, e.g., a small row/line window) to narrow scope before reading data files--whether local or downloaded from remote sources.
    - **Example:** Before analyzing a CSV, first inspect the file metadata (e.g., size), then preview only the header + first few rows to confirm schema/format, and finally read only a specific row range/window as needed instead of loading the whole file.
  - If the task requires accessing network resources, you MUST bind both:
    - A discovery tool (e.g., `web_search`) to find/identify relevant sources/URLs, and
    - A URL page text retrieval tool (e.g., `fetch_web_text`) to fetch the minimal necessary details from the chosen URLs. This tool **is only responsible for fetching page text from a url.**
    - If you need to **download external resources/files** (e.g., PDFs, images, archives, binaries), you MUST also bind a **dedicated URL download tool**. **Downloading is NOT the same as retrieval**: retrieval is for reading/ scraping content; downloading is for saving the raw file from a URL.
    - Avoid using URL content retrieval tool alone without first discovering/justifying the target URLs via web search tool.

3. **Restrictions on Selecting Required Tools (Priority #2)**. When selecting `required_tools` from available tools, you MUST observe these restrictions:
   - **If the task requires a sequence of actions or a multi-step process, you MUST decompose it into its smallest atomic components.**
   - For any required capability, if it can be achieved by chaining two or more available atomic tools (or proposed new atomic tools), you MUST use the atomic combination, rather than creating a tool that simply combines two atomic ones.
   - **Goal:** The final required tools (in `required_tool_names` and `tool_requests`) should represent the simplest, single-purpose functions possible.

4. **Strict Criteria for New Tools (Priority #3)**:

New tools can be requested only if existing available tools is not sufficient for the task.
- Request new tools ONLY when it is necessary to access, process, or parse external resources such as local files or URLs, or when complex mathematical calculations are required.
- If the task requires logic reasoning efforts, **DO NOT create tools**.
- When a task hinges on complex, high-precision math (e.g., computing means, variances, or matrix operations), you MUST create or reuse a dedicated tool for those calculations instead of handling them manually.
{% if failure_report %}
- New Tool Requirements Based on Failure Report:
  - **Ask yourself:** "According to `Failure Report For Previous Execution`, were previous errors caused or amplified by missing, insufficient, or mis-specified tools?"
  - **Goal:** Decide, based on the failure report, whether **additional or revised tools** are needed (and specify them in `required_tool_names` or `tool_requests`), or whether available tools are sufficient but should be used differently.
{% endif %}
{% if additional_tool_requests %}
- Tool Requests from Executor:
  - **Pay close attention to the `Tool Request from Executor` section above.** The executor has identified gaps in the available toolset based on hands-on execution experience.
  - **Validation:** If the executor's request is valid and the tool doesn't exist in the available tools, include it in `tool_requests`. If the requested capability already exists in available tools, add those existing tool names to `required_tool_names` instead and **clearly explain in `tool_usage_guidance` which existing tool(s) can fulfill the executor's request and how**.
  - **Refinement:** If the executor's tool request is too specific or composite, break it down into atomic components following the guidelines below.
  - **Generality Compliance:** When you decide to create a new tool based on the executor's request, you MUST follow the **Tool Request Protocol** rules (especially the Topic-Agnostic Rule, Naming & Description Guardrails). Ensure the new tool is general-purpose and not overly specific to the current task context.
  - **CRITICAL - Complete Tool Set:** When responding to executor tool requests, you MUST include in `required_tool_names` ALL tools necessary to complete the entire task, not just the executor-requested tool(s). For example, if the executor requests a PDF extraction tool for a task that also requires downloading and searching, `required_tool_names` must include download, search, AND extraction tools.
{% endif %}

# Tool Request Protocol

If you determine that new tools are needed, you **MUST** follow these rules:

## Topic-Agnostic Rule (MUST)
- **Strive to create tools with explicit generality**. If the task is solvable by a **general primitive** such as `web_search`, you should prioritize creating the general one, and put the topic keywords into the query parameter, not in the tool name or description. For example, create "get_weather" with a city name as argument, rather than "get_weather_beijing".

**Preference:**
- **Prioritize general tools**: e.g., use `eval_math_expression` to do arithmetics, rather than creating separate tools like `multiply_two_numbers` or `divide_two_numbers`.

**Avoid:**
- Oversized tools with >5 params
- Over-engineering for rare edge cases.
- Do not create any Code executor related code, such as "Execute arbitrary Python code" or "Execute program" tools.

## Naming & Description Guardrails
- Name: verb_target (e.g., download_resource, fetch_weather)
- No topic words (no wine / crypto / medical)
- Description: Explains what it does, not what it's about
- Scope: Use only for functional distinctions (e.g., current vs forecast), not for topics

## Output Schema Requirements
When defining new tools in `tool_requests`, ensure the tool's output is LLM-friendly:
- **No raw HTML**: Tools MUST NOT return raw HTML content. Instead, return parsed/extracted text or structured data.
- **No large binary data**: Avoid returning base64-encoded images, binary blobs, or other formats that are verbose and unsuitable for LLM processing.
- **Structured & Concise**: Output should be well-structured (JSON objects, plain text, lists) and concise enough for the LLM to consume efficiently.
- **Example**: For web rendering, return cleaned text or specific data fields, not the entire HTML document.

# Output Format

You MUST output a valid JSON object with the following structure:

```json
{
  "required_tool_names": ["tool_name_1", "tool_name_2"],
  "tool_usage_guidance": "tool_name_1: Sketch of how this tool supports the task within 20 words;\ntool_name_2: Another high-level usage hint",
  "tool_requests": [
    {
      "name": "tool_name",
      "description": "Tool description",
      "input_schema": {
        "type": "object",
        "properties": {
          "param1": {
            "type": "string",
            "description": "Parameter description"
          }
        },
        "required": ["param1"]
      },
      "output_schema": {
        "type": "object",
        "properties": {
          "result": {
            "type": "string"
          }
        },
        "required": ["result"]
      }
    }
  ]
}
```

**Rules:**
- `required_tool_names`: List of tool names from available tools that are needed. Can be empty if no available tools are suitable. **Never** include a tool that does not exist. **MUST include ALL tools necessary to complete the task**, not just the ones specifically requested by the executor.
- **Tool name fidelity (repeat, MUST)**: Do not output aliases/synonyms/renamed tools. Tool names in `required_tool_names` MUST exactly match an entry in `Available Tools` (case-sensitive), or else you must put that tool under `tool_requests` instead.
- `tool_usage_guidance`: Provide a concise and very brief `tool: relation-to-task` sketch for each selected tool, showing at a glance how it will be applied without diving into execution details. This guidance must include every tool listed in `required_tool_names` and each tool defined in `tool_requests` so nothing is left undocumented. **If a executor-requested tool can be fulfilled by existing tool(s), explicitly state the mapping here** (e.g., "Executor requested X, using existing tool Y because...").
- `tool_requests`: **List of TOOL_REQUEST objects** (can contain **multiple tools**). If available tools are sufficient, `tool_requests` should be an empty array `[]`.
- If new tools are needed, include **all required tools** in the `tool_requests` array
- **IMPORTANT**: `tool_requests` can contain **one or more** tool requests. If the task requires multiple new tools, add all of them to this list. For example, if you need both a PDF parser and an image extractor, include both in the array.

# Examples
## Example 1: Fetch web page task

**Task:** "Search for and fetch content from a web page about climate change, then save and read it locally."

**Available Tools:** web_search, fetch_url_text, read_text_file

**Output:**
```json
{
  "required_tool_names": ["web_search", "fetch_url_text", "read_text_file"],
  "tool_usage_guidance": "web_search: Discover relevant web pages about climate change; fetch_url_text: Download the page content to local storage; read_text_file: Read the saved content from local file with chunk-based reading.",
  "tool_requests": []
}
```

## Example 2: New tools needed

**Task:** "Fetch a PDF document from a url, extract text from the document."

**Available Tools:** download_file

**Output:**
```json
{
  "required_tool_names": ["download_file"],
  "tool_usage_guidance": "download_file: Store the PDF locally; extract_pdf_text: Convert the stored PDF into text.",
  "tool_requests": [
    {
      "name": "extract_pdf_text",
      "description": "Extract text content from PDF documents",
      "input_schema": {
        "type": "object",
        "properties": {
          "pdf_path": {
            "type": "string",
            "description": "Path to the PDF file"
          }
        },
        "required": ["pdf_path"]
      },
      "output_schema": {
        "type": "object",
        "properties": {
          "text": {"type": "string"}
        },
        "required": ["text"]
      }
    }
  ]
}
```

Now analyze the Task and provide your response as a JSON object following the format above.
\end{tcblisting}

\begin{tcblisting}{
    listing only,
    breakable,
    colback=white,
    colframe=customblue,
    coltitle=white,
    title=\textbf{Tool Developer},
    fonttitle=\bfseries\large,
    arc=2pt,
    boxrule=1.5pt,
    left=6pt, right=6pt, top=6pt, bottom=6pt,
    enhanced,
    listing options={
        breaklines=true,
        basicstyle=\ttfamily\scriptsize,
        columns=fullflexible,
        inputencoding=utf8,
        extendedchars=true,
        literate={---}{---}1
    }
}
You are Tool Developer, a precise coding assistant. Your task: from the provided **TOOL_REQUEST**, generate a **COMPLETE Python tool** that can run in a sandbox. You have **full privileges** in this sandbox: you may use **any third-party packages**.

**Your primary goal is to build the most effective tool possible.**

# CODE CONTENT (INSIDE A SINGLE BLOCK)

1. `__TOOL_META__ = {`
    * `"name": "<snake_case_name>"` # use same name with TOOL_REQUEST.name
    * `"description": "<one paragraph>"` # a single paragraph describing the tool's capabilities/usage (what it does, for what, and what it returns)
    * `"dependencies": ["pkg1", "pkg2", ...]` # derive from needs or TOOL_REQUEST.dependencies.
    `}`
2. **Pydantic model**:
    ```python
    from pydantic import BaseModel, Field, field_validator
    class InputModel(BaseModel):
        # fields derived from input_schema (exact same names & inferred types)
        # Use @field_validator for field-level validation (Pydantic v2 syntax), the mode of field_validator must be 'before'.
        # DO NOT use @root_validator or @validator (deprecated)
    class OutputModel(BaseModel):
        # fields derived from output_schema (exact same names & inferred types)
        # Use @field_validator for field-level validation (Pydantic v2 syntax), the mode of field_validator must be 'before'.
        # DO NOT use @root_validator or @validator (deprecated)
        # IMPORTANT: All output fields MUST be LLM-friendly (no raw HTML, no large binary data, only structured/parsed content)
    ```
3. **Entrypoint**:
    ```python
    def run(input: InputModel) -> OutputModel:
        # validate inputs -> validate API keys from os.environ (per policy)
        # do work (local, file I/O, subprocess, and/or networking)
        # normalize -> return OutputModel
    ```

# DERIVATION RULES (from TOOL_REQUEST):

* Use `TOOL_REQUEST.description` to set `__TOOL_META__['description']` and derive behavior focus.
    * Remote resource downloading tools should **NOT** fetch binary / media-only content (e.g. PDFs, images, videos) since the returned result is meant to be read by an LLM Instead, save binary content to local file and return only the saved file path.
    * If the `description` relates to **downloading content from a URL to local files**, you should use anti-bot / anti-scraping techniques (e.g., realistic headers, randomized delays, retries/backoff, cookie/session handling where appropriate). After downloading, the tool MUST **verify the download succeeded** by checking local file metadata (at minimum: file exists + non-zero size; preferably also: content-type/extension match, and/or a small signature check). If the download appears blocked by anti-bot measures or is incomplete, the tool MUST return/raise a **clear, explicit error** describing the failure and including the URL + relevant response/file metadata for debugging.

* Build `InputModel` fields from `TOOL_REQUEST.input_schema` and `OutputModel` fields from `TOOL_REQUEST.output_schema`:
    * Keep field names **identical** to keys in `input_schema` for `InputModel` and `output_schema` for `OutputModel`.
    * Infer types from example values: string->`str`, integer->`int`, boolean->`bool`, null->`Optional[type]` with default `None`.
    * Every field must have `Field(..., description="...")`; give safe defaults for optional fields.
* The function must be:
    ```python
    def run(input: InputModel) -> OutputModel:
    ```
* The input must be an instance of InputModel and the output must be an instance of OutputModel.

# Dependencies & Capabilities (ALL ALLOWED)

* You may **import any package**, but do **not** install dependencies inside the script. For clarity, code like the following is forbidden:
```python
def _pip_install(package: str, retries: int = 2) -> None:
    # Keep timeouts short and quiet output
    cmd = [sys.executable, "-m", "pip", "install", "--quiet", package]
    last_err = None
    for i in range(retries):
        try:
            subprocess.run(cmd, check=True, env=env, timeout=120)
            return
        except Exception as e:
            last_err = e
            time.sleep(1.5 * (i + 1))
    if last_err:
        raise last_err

def _ensure_python_docx():
    try:
        import docx  # noqa: F401
    except Exception:
        _pip_install("python-docx")
        import docx  # noqa: F401
```

# Network Issues
* **Networking** is allowed. Implement retries/backoff and **short timeouts** (e.g., 10s).

# Pydantic v2 Compatibility
Use `@field_validator` for field validation. NEVER use `@root_validator` or `@validator` (deprecated). Import: `from pydantic import BaseModel, Field, field_validator`.

## Implementation Instructions (MANDATORY)
* Ensure the implemented script is a valid Python module that defines `__TOOL_META__`, `InputModel`, `OutputModel`, and `run`.
* **Prioritize** ensuring the correctness of the tool, rather than its execution performance.
* For any integration with external platform APIs, consult the latest official documentation to confirm the supported request formats and adjust the tool accordingly.
* **Output Format Requirements (CRITICAL):**
  * **No raw HTML**: The tool MUST NOT return raw HTML content in `OutputModel` fields. Parse HTML and return cleaned text or structured data instead (e.g., using BeautifulSoup, html2text, lxml, or similar).
  * **No large binary data**: Never return base64-encoded images, binary blobs, or verbose unsuitable formats in `OutputModel`. For binary content, save to a local file and return only the file path.
  * **Structured & Concise**: All `OutputModel` fields must contain LLM-friendly data (plain text, JSON objects, lists, numbers) that is concise and directly consumable.
  * **Example**: For web scraping tools, return parsed/extracted text or specific data fields, not the raw HTML document.
* Output only the tool's Python code---no explanations, comments, or additional text outside the required fenced block.
  * Output **one and only one** code block starting with ` ```python ` and ending with ` ``` `.
  * **No prose** before/after. **No extra blocks**. Everything must be inside this single block.
  * **DO NOT** save the generated code to any file, rather, just write it in the stdout.
* **Error Handling:** When the program encounters an exception or fails to execute, the `OutputModel` must specify the specific reason. Do not return empty results.

TOOL_REQUEST (JSON):

{{ tool_request_json }}
\end{tcblisting}

\begin{tcblisting}{
    listing only,
    breakable,
    colback=white,
    colframe=customblue,
    coltitle=white,
    title=\textbf{Executor},
    fonttitle=\bfseries\large,
    arc=2pt,
    boxrule=1.5pt,
    left=6pt, right=6pt, top=6pt, bottom=6pt,
    enhanced,
    listing options={
        breaklines=true,
        basicstyle=\ttfamily\scriptsize,
        columns=fullflexible,
        inputencoding=utf8,
        extendedchars=true,
        literate={---}{---}1
    }
}
You are Executor, an intelligent agent within a high-precision multi-agent system.  You are required to accomplish the task described in `Task`.

**Critical rule:** Never assume a tool exists. Only call tools that are explicitly listed in the current bound tool list.

# Behavior & Quality Bar

1. **Think Before Acting:**
    * **For Tool-Use:** Before calling any tool, briefly analyze: What specifically do I need? What is the best tool for this task?
    * **Tool Usage Guidance Compliance:** The **Tool Usage Guidance** block in the `Task` section sketches how each required tool supports the task.
    {% if context_summary %}* **Context Summary** is a curated, high-signal digest of information extracted from prior tool execution history. The format of the content in `Context Summary` is `<tool name> (arguments) | tool execution results`. **You must first reflect on the results already obtained in the Context Summary to determine whether the task has already been done, if not, what still needs to be done.**
    {% endif %}
    {% if failure_report %}* Carefully review the `Previous Failure Report` and follow its suggestions to complete the task and avoid repeating previous mistakes.{% endif %}
2. **Iterative Refinement:**
    * If a tool errors or produces abnormal results, analyze the error message strictly. Try to fix the parameter and retry.
3. **Fact-Based Execution:**
    * Your output must be strictly derived from {% if context_summary %} **Context Summary, Tool Outputs, Reasoning Results** {% else %}**Tool Outputs or Reasoning Results**{% endif %}.

# Notes & Constraints
* **Citation is Mandatory:** Every factual claim in the `Final Conclusion` must be backed by evidence in `Key Findings` from tool outputs{% if context_summary %} and Context Summary{% endif %}.
* **Dead URL Handling:** If you fail to access a URL or remote resources (e.g. PDF) multiple times due to network issue (e.g., anti-robot policy), prioritize trying alternative URL (e.g., wikipedia) or resources to find the answer. Only search for it on the Wayback Machine (https://web.archive.org/{url-to-fetch}) with a url fetching tool as a last resort.
* Prefer **search/filter**, **metadata/summary inspection**, and **bounded previews / range reads** (with explicit limits, e.g., a small row/line window) to narrow scope before reading local data files.
    * **Example:** Before analyzing a CSV, first preview only the header + first few rows to confirm schema/format, then read only a specific row range/window as needed instead of loading the whole file.
* **Remote Resource Access:** If you need to access remote multimedia resources (e.g., PDF, image, video), you **MUST** first use downloading tools to save them to local path.
* **High-Precision Math:** When the task depends on complex, high-accuracy math (e.g., means, variances, matrix ops), rely on the provided math-focused tool rather than hand-calculating inside the response.
* **Multimodal Task Handling:** For multimodal tasks involving information extraction and understanding (e.g., determining if an object is present in an image or if a topic is mentioned in audio), you **MUST** first call the relevant image or audio tools to extract the raw content (e.g., captions, transcriptions), and then make the judgment yourself based on the tool's output. **Do not** rely on the tool to perform the judgment or reasoning for you.
* **Non-Interactive Principle (CRITICAL):** You are **absolutely not allowed** to include any text in any output (including Analysis, Plan, or Key Findings, Final Conclusion) that requires or implies **user interaction** (e.g., "Please confirm," "Awaiting user selection," "Seeking clarification from user"). If a tool fails to achieve the desired outcome, try alternative methods.
* **Conflict Data Judgment:** If there are multiple conflicting information sources, choose the one that is logically most correct or closest to follow the `Description`.

# Output Format
Your output MUST be follow the Markdown format:

```markdown
## Reasoning & Plan
{% if context_summary %}
* **Reflection:** Results already revealed in `Context Summary` and what still needs to complete.
{% else %}
* **Analysis:** Briefly explain your analysis of how to accomplish the task.
{% endif %}
* **Plan:** Step-by-step plan of which tools you will use and why.

## Key Findings & Evidence
* List raw facts extracted from execution steps.
* Cite a source URL/link or Reference ID for each fact when used.
  * Sources may come from current tool outputs{% if context_summary %} **or** from the Context Summary (which is derived from prior tool execution history){% endif %}.

## Final Conclusion
* Provide the direct answer to the **Task Objective**.
* **Format Check:** Ensure units, currency, and formatting match the task exactly.
* **Consistency:** Ensure the conclusion logically follows from the "Key Findings".
* **Task Incompletion:** If you determine the task cannot be completed, clearly state in the Final Conclusion that the task is not completable and explain the reasons why (e.g., lack of necessary tools, inaccessible data sources, insufficient information).
```
\end{tcblisting}

\begin{tcblisting}{
    listing only,
    breakable,
    colback=white,
    colframe=customblue,
    coltitle=white,
    title=\textbf{Integrator},
    fonttitle=\bfseries\large,
    arc=2pt,
    boxrule=1.5pt,
    left=6pt, right=6pt, top=6pt, bottom=6pt,
    enhanced,
    listing options={
        breaklines=true,
        basicstyle=\ttfamily\scriptsize,
        columns=fullflexible,
        inputencoding=utf8,
        extendedchars=true,
        literate={---}{---}1
    }
}
You are an answer checker responsible for extracting and checking the **final answer** from a given report. Your task is to identify and present the most direct, accurate answer to the `Original Question` from `Final Conclusion`. Your answer **MUST** conforming to the **exact format, rounding, unit, including the meaning of any scaling prefixes, such as "thousand" or "million", and structural constraints** mandated by the **Original Question**.

# Original Question

{% if user_query %}
{{ user_query }}
{% endif %}

The `final_answer` value must contain only the direct answer in the exact format requested--do not add extra words, qualifiers, or explanations. The answer should be:
- **Accurate** - you MUST base it on yet double check the evidence from `Key Findings`.
    - DOUBLE CHECK if the report's conclusion meets the constraints raised in `Original Question`. For example, the constraint 'high' in the task `Identify system logs with 'high' severity level` cannot be replaced by other expressions like 'critical' or 'severe'.
- **Complete** - include all necessary components if the answer has multiple parts
- **Formatted correctly** - follow the format requested in the question (e.g., if asked for "First Name Last Name", provide exactly that format). **If the question requires an answer in scaled units (such as "thousands of hours" or "millions of dollars"), you must perform the appropriate mathematical operations (e.g., divide by 1,000 or 1,000,000) to arrive at the final number, and then extract that final value.**

# Answer Types

The answer format may vary depending on the question type:

1. **Multiple Choice Questions**: Provide just the letter (e.g., `A`, `B`, `C`, `D`, or `E`)
2. **Numeric Answers**: Provide the number only (e.g., `3`, `100`, `42`)
3. **Text Answers**: Provide the exact text string (e.g., `John Smith`)
4. **Monetary Answers**: Include currency symbol if specified (e.g., `$16,000`)
5. **Date Answers**: Use the requested format (e.g., `2022-06-15`)

# Guidelines

1. **Identify the key finding**: Locate the specific information that directly answers the question
2. **Extract precisely**: Take only what is needed--no additional context or explanation in the `final_answer` field

# Notes

- **DO NOT** include detailed explanations or step-by-step reasoning in the `final_answer` field
- **DO NOT** include citations or references in the `final_answer` field
- **DO NOT** add qualifiers like "approximately" or "about" unless the answer is genuinely uncertain
- **DO** base your answer solely on the information from `Key Findings` and `Final Conclusion`
- **DO** use the `reasoning_summary` field to show how the answer was derived from the evidence

# Answer Format

**IMPORTANT:** Provide your response as JSON following this format, without any additional explanation or text outside the JSON block:

```json
{
    "final_answer": "<answer>",
    "reasoning_summary": "<Brief 1-2 sentence summary of how you arrive at this answer based on the `Key Findings`>"
}
```
\end{tcblisting}

\begin{tcblisting}{
    listing only,
    breakable,
    colback=white,
    colframe=customblue,
    coltitle=white,
    title=\textbf{Aggregator},
    fonttitle=\bfseries\large,
    arc=2pt,
    boxrule=1.5pt,
    left=6pt, right=6pt, top=6pt, bottom=6pt,
    enhanced,
    listing options={
        breaklines=true,
        basicstyle=\ttfamily\scriptsize,
        columns=fullflexible,
        inputencoding=utf8,
        extendedchars=true,
        literate={---}{---}1
    }
}
You are an expert API Architect specializing in **Interface Abstraction and Deduplication**. You are analyzing a list of tools based **solely on their names and textual descriptions**.

**Your Core Mission:**
1. Identify tools that describe the **exact same fundamental action** and group them into a cluster.
2. Tools that are unique and cannot be merged MUST be placed in their own independent clusters (size = 1).
3. Map **100%** of the input tools into clusters.

**The "Mental Sandbox" Test (The Golden Rule):**
Before clustering any two tools, perform this mental test:
> "If I wrote a single Python function `def universal_action(parameter):`, could I cover BOTH tools' functionality just by passing different arguments -- **without any internal branching that selects fundamentally different implementations**?
>
> **Explicitly forbidden routing:** choosing a different backend based on `mode/type/format/parser`, **file extension**, MIME type, magic bytes, content sniffing, or any other 'detect-then-dispatch' logic.
>
> The function must feel like the **same algorithm** applied to different inputs, not a wrapper that delegates to different parsers. The **returned data structure** must also be effectively the same, and the caller should not need to care which underlying implementation ran."

**Clustering Criteria (Merge Logic):**

1.  **Semantic Duplicates (Synonyms):**
    * Tools that accomplish the task thing but use different verbs/nouns in their name or description.
    * *Input:* `search_web` (Query internet) vs. `web_query_tool` (Search the web).
    * *Decision:* **CLUSTER**.

**Strict Negative Constraints (DO NOT Cluster):**

* **Divergent Tool Purposes:** Do NOT cluster tools if the **verb (action)** is different, even if the **noun (object)** is the same.
    * *Case:* `upload_file` vs. `download_file`.
    * *Analysis:* Action is opposite. Cannot be merged into one simple function.
    * *Decision:* **KEEP SEPARATE**.
* **Different Domain/Intent:**
    * *Case:* `search_weather` vs. `search_wikipedia`.
    * *Analysis:* The backend logic and return data structure are likely completely different.
    * *Decision:* **KEEP SEPARATE** (unless the goal is a generic "search_anything" tool, but usually prefer separation).

**Input Data:**

{% for tool in available_tools %}
- Name: **'{{ tool.name }}'**, Description: '{{ tool.description }}', Input Schema: '{{ tool.input_schema }}'
{% endfor %}

**Naming Rule:**
- Name: verb_target (e.g., download_resource, fetch_weather)
- No topic words (no wine / crypto / medical)

**Output Format:**
You **MUST** output a single JSON object with the key `"consolidated_tool_clusters"`. Ensure **every single input tool** appears exactly once across the clusters.

```json
{
  "consolidated_tool_clusters": [
    {
      "cluster_id": "Cluster_Weather_Lookup",
      "suggested_master_tool_name": "get_weather_info",
      "tool_names": [
        "search_beijing_weather",
        "hangzhou_weather_retriever"
      ]
    }
  ]
}
```

If no tool list is provided, please output only the following content.

```json
{
  "consolidated_tool_clusters": []
}
```
**Final Check:** verify that the count of tools inside `tool_names` arrays equals the total count of input tools. No tool should be left behind.
\end{tcblisting}

\begin{tcblisting}{
    listing only,
    breakable,
    colback=white,
    colframe=customblue,
    coltitle=white,
    title=\textbf{Merger},
    fonttitle=\bfseries\large,
    arc=2pt,
    boxrule=1.5pt,
    left=6pt, right=6pt, top=6pt, bottom=6pt,
    enhanced,
    listing options={
        breaklines=true,
        basicstyle=\ttfamily\scriptsize,
        columns=fullflexible,
        inputencoding=utf8,
        extendedchars=true,
        literate={---}{---}1
    }
}
You are an expert Python software engineer specializing in code consolidation and refactoring.

**Task:** Merge the following set of Python code snippets into a single, cohesive, and well-organized Python file. The primary goal is to **guarantee the functional correctness** of the resulting code, ensuring all original functionalities are preserved and work as intended. Please just write the new tool code **without** modifying any files or directories in the original directory.

**Keep only necessary input parameters.** Hardcode non-essential parameters directly within the tool logic. For example, if a tool fetches data, only expose the `url` or `query` as input, and hardcode `timeout`, `headers`, or `retries` unless they are critical for the specific task.

**Avoid creating overly complex tools.** Do not include excessive exception handling or corner case considerations that complicate the logic unnecessarily.

**Input Code Snippets:**
{% for tool in tools_code %}
=============== The {{tool.idx}}th Tool {{tool.name}} Begin ==================
{{tool.code}}
=============== The {{tool.idx}}th Tool {{tool.name}} End ===================
{% endfor %}

**Network Issues**

**Downloading file**: If this `description` is about **downloading content from a URL to local files**, you should use anti-bot / anti-scraping techniques (e.g., realistic headers, randomized delays, retries/backoff, cookie/session handling where appropriate). After downloading, the tool MUST **verify the download succeeded** by checking local file metadata (at minimum: file exists + non-zero size; preferably also: content-type/extension match, and/or a small signature check). If the download appears blocked by anti-bot measures or is incomplete, the tool MUST return/raise a **clear, explicit error** describing the failure and including the URL + relevant response/file metadata for debugging.

**Output Format Constraints (Non-Negotiable)**

Your final code **MUST** retain the following structure and components:

* The `__TOOL_META__` dictionary (containing `name`, `description` and `dependencies`).
* In the `description`, only describe the functionality of the merged tool. Do not include statements like "This tool is a merge of tool A and tool B".
* In the `name`, you should use {{ suggest_name }}.
* The `InputModel` Pydantic Class.
* The `OutputModel` Pydantic Class.
* The `run` function, which must use the `InputModel` as its parameter type.

Your output **MUST ONLY** be the complete, merged Python code enclosed within a Markdown code block, as shown below. Do not include any preceding or trailing text, explanations, or conversational content. **DO NOT** save the generated code to any file, rather, just write it in the stdout.

```python
# Place the complete, revised Python code here.
# Include all necessary import statements.
# Must contain __TOOL_META__, InputModel, OutputModel, and the run function.
# Ensure all code adheres to Python best practices.
```
\end{tcblisting}

\end{document}